\documentclass[11pt]{article}
\usepackage[a4paper]{geometry}
\usepackage[utf8]{inputenc}
\usepackage[T1]{fontenc}

\usepackage{authblk}
\usepackage{graphicx}
\usepackage[table]{xcolor}
\usepackage[english]{babel}
\usepackage[hidelinks]{hyperref}
\PassOptionsToPackage{hyphens}{url}\usepackage{hyperref}
\usepackage{subcaption}
\usepackage[final]{changes}
\usepackage{longtable}
\usepackage{multirow}
\usepackage{tipa}
\usepackage{csquotes}
\usepackage[
    style=authoryear,
    sorting=nyt,
    maxcitenames=2,
    maxbibnames=999,
    uniquename=false,
    uniquelist=false
    ]{biblatex}
\title{Over-representation of phonological features in basic vocabulary doesn't replicate when controlling for spatial and phylogenetic effects}
\author[1,2,*]{Frederic Blum}
\affil[1]{Department of Linguistic and Cultural Evolution, Max-Planck Institute for Evolutionary Anthropology, 04103 Leipzig}
\affil[2]{University of Passau, Chair of Multilingual Computational Linguistics, Passau, 94032, Germany}
\affil[*]{\href{mailto:frederic\_blum@eva.mpg.de}{Corresponding author: frederic\_blum@eva.mpg.de}}

\begin{document}
\maketitle

\begin{abstract}
    The statistical over-representation of phonological features in the basic vocabulary of languages is often interpreted as reflecting potentially universal sound symbolic patterns.
    However, most of those results have not been tested explicitly for reproducibility and might be prone to biases in the study samples or models. Many studies on the topic do not adequately control for genealogical and areal dependencies between sampled languages, casting doubts on the robustness of the results.
    In this study, we test the robustness of a recent study on sound symbolism of basic vocabulary concepts \replaced{which analyzed}{in} 245 languages.\deleted{Since} The new sample includes data \replaced{on 2864 languages from Lexibank}{from more than 2,800 languages}. We modify the original model by adding statistical controls for spatial and phylogenetic dependencies between languages.
    The new results show that most of the previously observed patterns are not robust, and in fact many patterns disappear completely when adding the genealogical and areal controls. A small number of patterns, however, emerges as highly stable even with the new sample.
    Through the new analysis, we are able to assess the distribution of sound symbolism on a larger scale than previously. The study further highlights the need for testing all universal claims on language for robustness on various levels.
\end{abstract}

\clearpage
\section{Introduction}
The arbitrariness of the sign is one of the most common fundamental concepts taught in linguistics \parencite{Saussure1916}. Yet, this concept has come under criticism as statistical evidence for sound symbolic patterns has piled up \parencite{nuckolls1999case, Blasi2016, ErbenJohansson2020, Cwiek2024}, and arguments for the role of iconicity during language evolution have been brought forward \parencite{Dingemanse2015, Dingemanse2018, Cwiek2021}. However, few of the quantitative studies are replicated on a large scale. Many studies revisiting previous publications for other topics in linguistic typology have found diverging results from previously established claims \parencite{Hartmann2024a, Becker2025}. This is a potential problem for inferences about universal cognitive structures related to the question of sound symbolism, since we do not have an assessment of those results on a world-wide level. The question emerges whether previous observations of the statistical over-representation of phonological features in basic vocabulary are robust, or whether they are, for example, an artifact of inherent biases in the sample or the methodology employed. Considering the availability of large-scale datasets for lexical data, we are now in a position to systematically compare the results from previous studies with results based on new datasets and improved statistical methods. Here, we present a study that looks at the results from \textcite{ErbenJohansson2020}, who tested 340 concepts of basic vocabulary in 245 languages for a statistical over-representation of phonological features. We expand their sample of languages to more than 2,800 of the world's languages in order to have a broad assessment of the robustness of their results.


The systematic test whether scientific results hold up under new circumstances (e.g. annotation choices, sample, statistical methods) is often called `replication' or `robustness analysis'. The general idea is that a replication tries to mimic a previous study as closely as possible, for example by running the same model on new data, while a `robustness analysis' contains a broader range of studies and is more free to make changes compared to the previous study \parencite{Goodman2016}. While a replication always tests the robustness of a result, it is possible to test the robustness without a conducting a replication study. But, to use the wording of \textcite[4]{Goodman2016}, `whether a study design is similar enough to the original to be considered a replication, a “robustness test,” or some of many variations of pure replication that have been identified, \textelp{} is an unsettled question'. While some scholars opt for a broad definition of `replication' where changes to the model are acceptable or even desired \parencite{Becker2025}, other opt for a narrow definition where this is not possible \parencite{Clemens2017}. In order to be explicit about what our study is, we adopt the definitions proposed by \textcite{Goodman2016} and use the term `results reproducibility' instead. This term is defined as `the conduct of an independent study whose procedures are as closely matched to the original experiment as possible' \parencite[2]{Goodman2016}. This definition avoids the different expectations that are created when the term `replication' is used. While this definition bears close resemblance to a broad definition of `replication', it explicitly allows changes to the model structure where necessary. We think of the addition of genealogical and areal controls as exemplary reasons for such necessary changes.

Closely related to the question of robustness is the issue of generalizability \parencite{Yarkoni2020}. Generalizability is related to the possibility of generalizing results beyond the sample to make claims about statistical universality, a common goal in linguistic typology \parencite{Bickel2011, Winter2021}. In studies of sound symbolism, for example, the geographical location of languages could impact their phonological features through areal spread or undetected borrowings. Similarly, genealogical relationships between languages can introduce dependencies that skew the results. If a specific rare feature is present in a large language family but not in other families, it is very likely that this presence is due to inheritance rather than broader evolutionary patterns. Both genealogy and areal relationships are examples where certain results can be biased by the sample, which causes problems for a generalization beyond the sample to a new population.

The lack of robustness is a general problem in quantitative research and relates to the fact that many established research results do not hold up when tested on new data \parencite{Ioannidis2005, OSC2015}. While the issue of robustness has found some attention in quantitative typology \parencite{Corbett2005, Bisang2011, Hartmann2024a}, systematic approaches are scarce until now \parencite{Editors2006, Song2007, Becker2025}. For individual findings,\deleted{however,} studies revisiting new data and additional controls on the same research question have found severe problems on specific topics. As \textcite{Becker2025} summarize, there have been multiple studies in which adding controls for phylogeny and contact have contradicted previous claims of strong effects. The most well known case is probably that of a relation between financial savings and the presence of a grammatical future tense in a language \parencite{Chen2013}. Adding phylogenetic controls for the genealogical relationships between languages made the effects disappear \parencite{Roberts2015}. Similar claims and rebuttals have been reported for the diversity of phoneme inventories \parencite{Atkinson2011, Jaeger2011} or the relation of pitch and humidity \parencite{Everett2015, Hartmann2024a}. Following the conclusion of \textcite{Becker2025}, we see that it is worthwhile to include a broad sample of languages, and to control for the dependencies between data-points statistically \parencite{Winter2021}. A key issue that emerged from those studies, and which future studies need to have in mind, is the necessity of accounting for areal and genealogical dependencies between data points to avoid statistical fallacies \parencite{GuzmanNaranjo2021}. 

In other areas of quantitative linguistics, like psycholinguistics or corpus analysis, the issue of dependencies between data points has often been tackled with maximal study designs \parencite{Jaeger2011, Barr2013}. In quantitative typology,\deleted{however,} scholars often prefer to hand-pick `balanced' samples \parencite{Bakker2010}. However, this reduces the amount of data that is provided to a model and might create unwanted sample biases. As recent studies point out, a much better way to counter areal and genealogical biases is through statistical control \parencite{GuzmanNaranjo2021, Becker2025}. This provides a more complete picture of the complex reality of linguistic phenomena, since we can include data from more languages and design a statistical model that accounts for the non-independence between data points. By applying those methods, researchers can improve the reliability and validity of their findings, ultimately advancing our understanding of linguistic phenomena and their implications in typological studies.

The paper offers three main contributions. First, it presents a systematic comparison of previous findings of sound symbolic patterns to a new analysis based on a large sample of languages, allowing scholars to have an independent assessment of the robustness of \added{the }previously discovered patterns. Apart from the added controls for genealogical and areal dependencies between data points, we use the exact same model structure as the original study \parencite{ErbenJohansson2020}. This includes both the phonological annotation schema with its different categories as well as the model family itself. We therefore frame this as testing the `results reproducibility' \parencite{Goodman2016}. Second, our analysis allows us to discuss the implications of the combined results for overarching topics such as the relevance of sound symbolism for historical linguistics and the link of sound symbolism with cognition, language, and language evolution. Third, we showcase how large-scale databases can be used for robustness analysis in quantitative linguistics while taking statistical biases into account, \replaced{which}{and} opens pathways for further, similar studies.

\section{Background}
The relation between form and meaning is not completely arbitrary, but it is not deterministic either \parencite{Saussure1916, Hockett1960, nuckolls1999case, Dingemanse2015}. The non-arbitrary presence of certain phonological features in vocabulary is called sound symbolism. The general idea is that sound symbolism reflects cognitive biases in language, with implications for speech acquisition and processing \parencite{Perniss2014, Massaro2017, Tamariz2017}. In this context, the study of sound symbolism offers valuable insights into the relationship between sound, meaning, and cognition. By examining how specific phonological features are associated with certain meanings across various languages and language families, linguists can gain a better understanding of how languages have evolved over time and how sound symbolism may have played a role in the emergence of language \parencite{ErbenJohansson2020}. As such, iconic gestures and sound symbolism are argued to provide one of the directly observable interfaces between cognition and language \parencite{Dingemanse2018, DiPaola2024}. For example, it has been shown that a strong relation between visual perception and speech sounds exists across cultures \parencite{Margiotoudi2020, Cwiek2021}. The analysis of such results \added{that are believed to be universal} across non-WEIRD languages is a necessary component of establishing strong links between language and cognition \parencite{Henrich2010, Evans2009, Blasi2022, Blum2024c}.

Two studies stand out with respect to the cross-linguistic investigation of sound symbolic patterns on a large scale. First, \textcite{Blasi2016} analyzed sound symbolic patterns in 40 items of basic vocabulary across more than 6,000 languages. In contrast to later studies, the authors investigated the positive and negative association of symbols within the simplified ASJP transcription system of 41 phones (\cite{ASJPv20}, \url{https://clts.clld.org/contributions/asjp}) \replaced{analysing}{with} 40 concepts of core vocabulary. Across their dataset, they found 74 statistically significant sound-meaning associations over 30 different concepts and 23 sound symbols. This means that they found apparent statistical over-representations of certain sounds in 75\% of \replaced{the analysed concepts}{a list of core vocabulary}. Even though the study was not the first to analyze the cross-linguistic distribution of sound symbolism, it has, until today, the largest sample size of languages. They ran individual models per macro-area, and considered statistical over-representations as significant if multiple macro-areas had an observed statistical effect beyond a certain significance threshold.
 
Second, \textcite{ErbenJohansson2020} analyzed the sound-meaning associations in 245 languages across a list of 344 concepts of basic vocabulary. The larger amount of concepts comes at the expense of having data from less languages available. Instead of a simplified phonological inventory, this study was based on phonological features (e.g. voicing, place of articulation, manner of articulation, vowel height). This model design allows for a more fine-grained analysis of the sound symbolic patterns in question. One of the main claims of \textcite{ErbenJohansson2020} is that sound symbolism is more widespread among basic vocabulary than previously assumed. Depending on the semantic analysis, they report between 25\% and 50\% of concepts in basic vocabulary lists like Swadesh-100 \url{https://concepticon.clld.org/contributions/Swadesh-1955-100}, Leipzig-Jakarta (\url{https://concepticon.clld.org/contributions/Tadmor-2009-100}), and the ASJP list (\url{https://concepticon.clld.org/contributions/Holman-2008-40}) being affected by patterns of sound symbolism. Based on those results, the authors argue that the prevalence of sound symbolism in basic vocabulary raises doubts about the validity of such lists to assess the genealogical relationship between languages. The authors also present a detailed typological classification of sound symbolic patterns based on their findings and relate those to different processes of language evolution, on which we will not expand here. Instead, we focus on the first research question of the original study about the cross-linguistic extent of the statistical over-representation of sounds in sound-meaning associations.

Another key contribution by \textcite{ErbenJohansson2020} is an explicit comparison between the results of their study and \textcite{Blasi2016}, making it possible to see the overlap and divergences between both results. This is already a first test of robustness, showing that many patterns of \textcite{Blasi2016} could not be found by \textcite{ErbenJohansson2020}. Other patterns however, like a nasal feature for pronouns or a lateral feature for \textsc{tongue} have been found in both studies. No quantitative perspective on the amount of robust patterns has been provided, so it is difficult to estimate how many features are robust, and how many should be considered unstable. However, the tabular comparison makes it possible to proceed in a qualitative way.

In summary, one can say that both studies have different strengths: While the first study is based on the analysis of a large amount of languages, the second study offers the more detailed phonological analysis over a greater amount of concepts, albeit on fewer languages. The goal of the present study is to combine both approaches, and to test the analysis of \textcite{ErbenJohansson2020} for robustness across a large sample of languages, while statistically accounting for the areal and genealogical non-independence of data points. We conclude with a comparison between robust concepts from all three studies.

\section{Methods}
The present study intends to tackle the issue of results reproducibility by re-running the analysis of \textcite{ErbenJohansson2020} on a larger sample of languages, while accounting for statistical issues of generalizability in form of areal and genealogical relations. Thankfully, the authors also share their code and data, making it easy to follow the original analysis and investigate their statistical model in detail. In order to avoid genealogical dependencies between languages, they restricted their sample to one language per language family. Here, we take the approach to use as much available high-quality data as possible, in order to better generalize beyond the sample. This follows recent investigations that have shown that sampling as bias control is not ideal, and that we should rather implement statistical controls for phylogenetic and areal effects \parencite[37]{Becker2025}. Through the addition of statistical bias controls for phylogeny and areal contact, we try to exclude potential confounders and have a better assessment of the spread and characteristics of over-represented phonological features across concepts of basic vocabulary in the languages of the world.

\subsection{Data}
The data for this study is taken from Lexibank \parencite{List2022e, Blum2025}, a large-scale database for comparative wordlists. The most recent release (v2.1) contains data from more than 3,000 languages. The data is standardized for comparability and inter-operability on three different levels. The datasets that form part of Lexibank have standardized transcriptions that adhere to the Cross-Linguistic Transcription Systems (CLTS, \cite{CLTS}). For this purpose, all datasets have individual orthography profiles, which map the graphemes to a CLTS-conform phone. The languages are mapped to unique identifiers in Glottolog \parencite{Glottolog}. The lexical concepts are mapped to Concepticon, a repository of basic vocabulary lists \parencite{Concepticon}. Since both Lexibank and Concepticon adhere to the Cross-Linguistic Data Formats \parencite{Forkel2018}, it is easy to combine the use of both and query the Lexibank data for specific concepts. Of the 344 concepts from the original study, 284 are mapped to Concepticon and will be used for the our study (\url{https://concepticon.clld.org/contributions/Johansson-2020-344}). The remaining concepts are mainly kinship terms from specific gender perspectives, which do not have a fine-grained representation in Concepticon. They are neither part of the basic vocabulary lists nor were they reported to be highly sound-symbolic in the original analysis. In order to have the automated retrieval of data based on Concepticon mappings, we argue that omitting of these concepts is necessary.

Furthermore, there were some inconsistencies in the concept mappings in Concepticon between the Johansson-2020-344 list (\url{https://concepticon.clld.org/contributions/Johansson-2020-344}) and the basic vocabulary lists. Since the original article claims to feature all the concepts present in Tadmor-100, Swadesh-100, and Holman-40, we included all concepts that are mapped to Concepticon but not in Johansson-2020-344. For example, Johansson-2020-344 features \textsc{we (inclusive)} (\url{https://concepticon.clld.org/parameters/1131}) and \textsc{we (exclusive)} (\url{https://concepticon.clld.org/parameters/1130}), but not \textsc{we} (\url{https://concepticon.clld.org/parameters/1212}), which is the concept representation in Holman-2008-40 list (\url{https://concepticon.clld.org/contributions/Holman-2008-40}). This highlights both the advantages and disadvantages of using such an automated retrieval. On the one hand, it is semantically explicit. Only those concepts are retrieved that are explicitly called for. On the other hand, the \textsc{we}-example shows the problems with studies that did not use such semantic mappings in the beginning \parencite{List2016}, where some relevant concepts might be omitted. Instead of making such an approach problematic, from our point of view this step shows the necessity of using such concept mappings in order to disambiguate the semantics of different concepts. When working with global samples, a manual annotation is not feasible. And for working with sound symbolism, it is indispensable to distinguish\deleted{ between}, for example, \textsc{blow (of wind)} from \textsc{blow (with mouth}), where only the latter tends to be associated with \replaced{sound symbolism in the form of rounded vowels}{the human body}.

We accessed the data through a SQLite query that filters the Lexibank database for concepts that are used in the original study. In the query, we also compute the phonological features of all segments, based on their CLTS string value \parencite{CLTS}. For example, the phoneme /\textipa{tS}/ has the CLTS string `voiceless post-alveolar sibilant affricate consonant'. The query converts those features \added{into} the ten phonological categories from the original study. The standardization provided through CLTS makes it possible to have a quick, efficient, and also transparent way of annotating the phonological features for standardized datasets published in Lexibank \parencite{Blum2025}. The individual mapping categories can be retrieved from the query that is published within the GitHub repository \added{for this study linked in the Supplementary Material. The most frequent vowels and consonants in the data are presented in Table~\ref{tab:freq_vowels} and Table~\ref{tab:freq_cons} together with their phonological features.}

\begin{table}
    \centering
    \begin{tabular}{|c|c|c|c|c|c|c|}\hline
    \textbf{IPA} & \textbf{CLTS} & \textbf{rounded} & \textbf{back} & \textbf{height} & \textbf{extreme} & \textbf{count} \\ \hline
    a & unrounded open front & unrounded & front & low & low-front & 276102 \\\hline
    i & unrounded close front & unrounded & front & high & high-front & 159648 \\\hline
    u & rounded close back & rounded & back & high & high-back & 128213 \\\hline
    o & rounded close-mid back & rounded & back & mid & high-back & 89114 \\\hline
    e & unrounded close-mid front & unrounded & front & mid & high-front & 88414 \\\hline
    \end{tabular}
    \caption{Phonological features for the five most frequent vowels.}
    \label{tab:freq_vowels}
\end{table}

\begin{table}
    \centering
    \begin{tabular}{|c|c|c|c|c|c|c|}\hline
    \textbf{IPA} & \textbf{CLTS} & \textbf{voicing} & \textbf{manner} & \textbf{position} & \textbf{count} \\\hline
    n & voiced alveolar nasal & voiced & nasal & alveolar & 111152 \\\hline
    m & voiced bilabial nasal & voiced & nasal & labial & 89748 \\\hline
    k & voiceless velar stop & unvoiced & stop & velar & 89295 \\\hline
    t & voiceless alveolar stop & unvoiced & stop & alveolar & 75441 \\\hline
    l & voiced alveolar lateral approximant & voiced & lateral & alveolar & 68321 \\\hline
    \end{tabular}
    \caption{Phonological features for the five most frequent consonants.}
    \label{tab:freq_cons}
\end{table}

We have used data from all languages in the Lexibank release that were not part of the initial study, in order to test the robustness on a whole new set of languages. After removing the languages that were part of the original study based on glottocodes, data from 2,862 languages remains in the dataset. In total, the filtered dataset consists of 2,009,692 phones from 419,494 lexical forms. Of those, around 46.6\% are vowels, and 53.4\% are consonants.

\subsection{Model}
The original study presents a Bayesian linear regression analysis using a Dirichlet model for the individual phonological categories. Instead of sampling only one language from each language family, we opt for a full sample of the available data in Lexibank and expand the original model by adding a statistical bias control for genealogy and contact. For this purpose, we add two multilevel intercepts based on covariance matrices for phylogenetic and areal distances. For the phylogenetic regression, we follow the approach by \textcite{Becker2025}. The covariance matrix represents the phylogenetic relationship between languages in a gradual way, thereby reflecting the assumption that we expect a stronger correlation between more closely related languages than with distantly related ones. The matrix consists of a standardized measure of the path length between two languages of the same family in the Glottolog tree. As a proxy for spatial covariance, we compute the geodesic distances based on the Glottolog coordinates for all languages in the same macro-area using the geopy package \parencite{GuzmanNaranjo2023, geopy}. Distances larger than 1,000km are set to zero in the covariance matrix, assuming that those languages have not been in contact. Both covariance matrices are created while retrieving the data from Lexibank through the before-mentioned SQLite query. While both distance measures are not perfect representations of the true world, we believe that they represent reasonable approximations that fulfill their purpose as statistical controls \parencite{GuzmanNaranjo2021, GuzmanNaranjo2022}. 

Following the methodology of the original study, we run a Dirichlet model for all target features with varying effects for word and language. We use the same 10 levels of phonological features than the original study, five for consonants and five for vowels. For consonants, we analyze voicing (two levels), position (five levels), manner (five levels), and the combination of manner and position with voicing (each ten possible levels). For vowels, we analyze rounding (two levels), height (three levels), backness (three levels), \added{the} combination of height and backness (four levels), and the combination of vowel height, backness, and rounding (eight levels). \added{The individual levels are presented in Table~\ref{tab:vowels} for the vowels and in Table~\ref{tab:consonants}, both in Supplementary Material D.}
 
\subsection{Evaluation thresholds}
We evaluate the \replaced{results}{model} by running posterior predictive simulations \parencite{Gelman2013, McElreath2020, Heiss2021}. For those posterior predictions, we evaluate the Highest Posterior Density Intervals (HPDI) for the proportion of phonological features. Following the description in the original study, we use the natural logarithm of 1.25 (upper threshold, 25\% increased odds ratio for a specific feature) and of 1/1.25 (lower threshold, 25\% reduced odds ratio for a specific feature) as thresholds to evaluate the log odds ratio of each pattern. The area between those thresholds is considered the Region Of Practical Equivalence (ROPE, \cite{Kruschke2018}) to zero. \replaced{Effects in this area are considered}{that is, effects that are} so small they are basically indistinguishable from zero. We also use the exact same levels as in the original study for defining the strength of any emerging effects:

\begin{figure}[ht]
    \begin{enumerate}
        \item[Strong effect:] The 95\% Highest Posterior Density Interval (HPDI) is completely outside the ROPE.
        \item[Weak effect:] The mean is outside the ROPE, and the 95\% HPDI fully positive or negative.
        \item[No effect:] The 95\% HPDI is completely within the ROPE.
        \item[Not interpretable:] Neither condition is fulfilled.
    \end{enumerate}
\end{figure}

\subsection{Prior distributions}
\begin{figure}[ht]
    \centering
    \includegraphics[width=\linewidth]{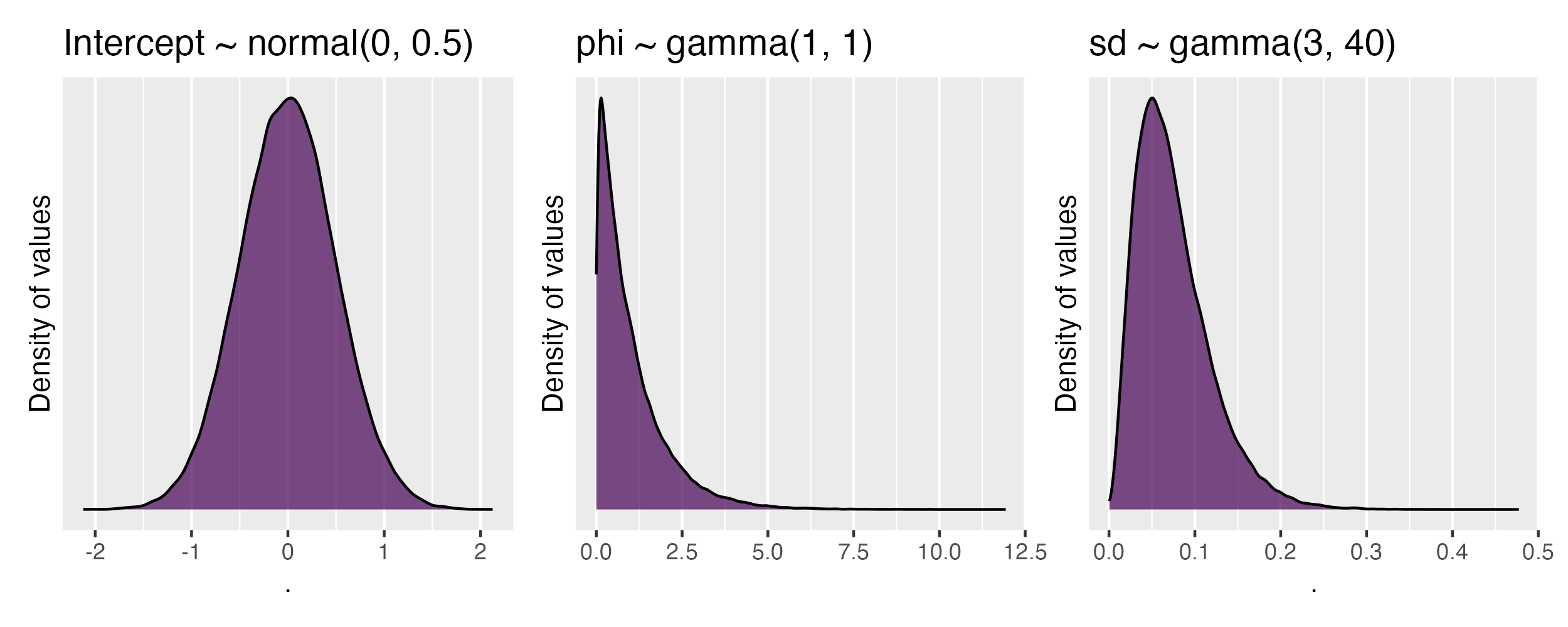}
    \caption{Prior distributions for all model parameters.}
    \label{fig:priors}
\end{figure}

We have updated the model with weakly informative prior distributions instead of brms default priors from the original study \parencite{Gelman2013, McElreath2020}. This is necessary due to the addition of the statistical bias controls, which make the model harder to converge. Following standard practices of Bayesian modeling, we run prior simulations before running the final model \parencite{Gelman2013, Kruschke2018, McElreath2020}. We do this for all model parameters to make sure that the \replaced{priors are}{model is} a) not influencing the results, and b) making reasonable assumptions about the order of magnitudes of any effects. We can base this prior knowledge on the results of \textcite{ErbenJohansson2020} and the reported effect size of the statistical over-representation of sounds. The results of the prior simulation are presented in Figure~\ref{fig:priors}.

Most importantly, the intercept of each feature is modeled with a mean of 0. With the selected priors, we do not expect the intercept of any effect to be much bigger than 1 (on log-scale). For comparison, the highest mean for any intercept in the original study was around 0,5. Thus, our prior distributions are considered `weakly informative'. The gamma distribution for $\phi$ (`phi') provides the prior distribution for the areal and phylogenetic covariance matrices. As one can see from the distribution, the basic assumption is that the effect of spatial and phylogenetic covariance becomes weaker with increasing distance in the matrix, eventually reaching zero \parencite{McElreath2020, GuzmanNaranjo2021}.

\subsection{Implementation}
The preprocessing of the Lexibank data is done with Python in combination with \emph{SQLite} \parencite{sqlite}, following the example from \textcite{Blum2024c}. Through the provided segmentation and standardization of the data in Lexibank, this preprocessing comes in a straightforward manner. The models are implemented in \emph{R} \parencite{R_4.4.3} using \emph{brms} \parencite{brms} with a \emph{cmdstanr} backend \parencite{cmdstanr}. For simulating the posterior draws, we use the \emph{tidybayes} package \parencite{tidybayes}. The results are visualized using \emph{ggplot2} \parencite{ggplot}. \added{All code used for this study is curated on GitHub (+++link removed for anonymity+++). For the review, the code can be accessed anonymously on OSF: \url{https://osf.io/4kg52/?view_only=a96702c55db14528b9a3e7ed3701588b}}.

\section{Results}
\subsection{Estimating the posterior distribution of the new sample}
\added{Following the setup from the original study}, we run individual models for all ten phonological categories\deleted{that have been analyzed in the original study}. Based on the posterior distributions of each model, we simulate the expected draws for all concepts and compute the log-likelihood of individual phonological features. We then compare the log-likelihood to our thresholds to establish which effect can be considered strong, weak, or absent. Finally, we plot the results against the previous observations. A detailed overview of the individual effects for all ten phonological categories are presented in the Supplementary Material A.

\subsection{Overall comparison between previous and new results}
\begin{figure}[ht]
    \centering
    \includegraphics[width=0.9\linewidth]{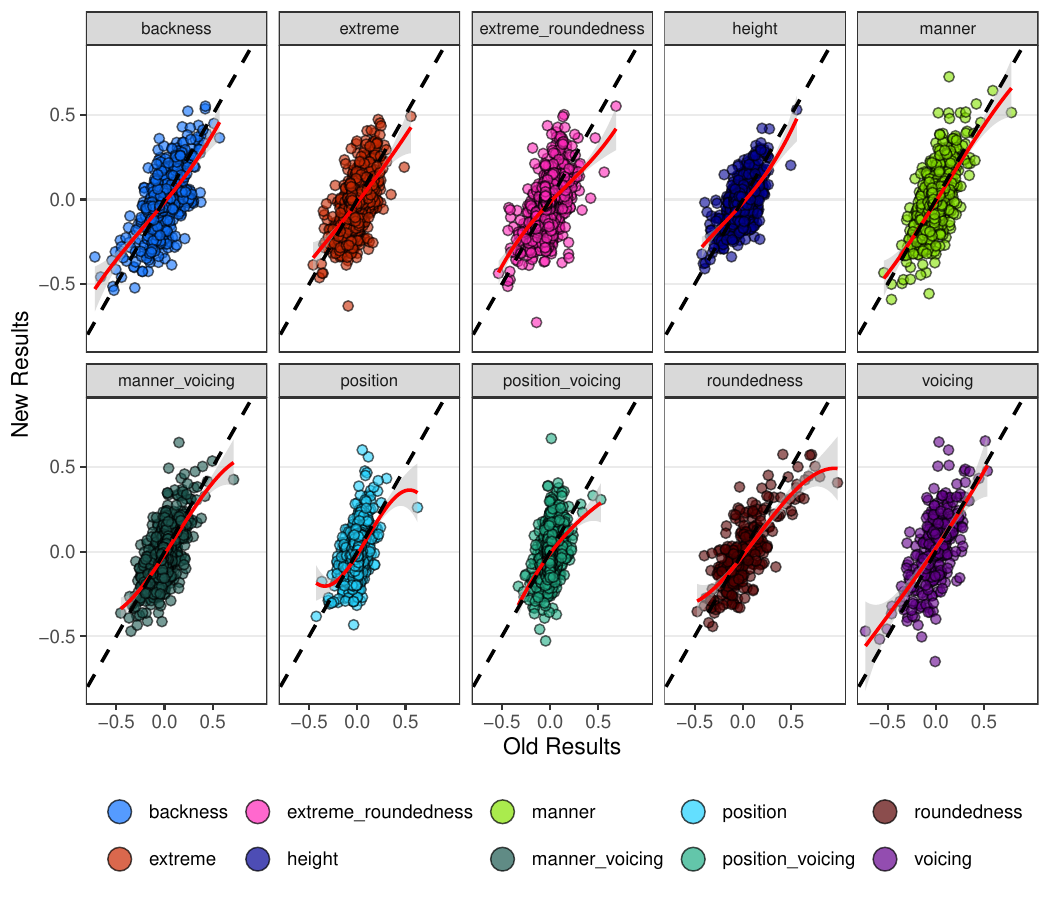}
    \caption{Correlation scatter plot between old and new models for all concepts and their phonological features. The mean values for all phonological features across all concepts are directly compared with each other. The dashed line indicates a perfect correlation, whereas the red line represents a fitted linear model with three cubic B-splines.}
    \label{fig:corr}
\end{figure}

A direct comparison of previous and new results shows that while there is some correlation for all combinations of concepts and phonological features (Pearson's \textit{r} 0.6), the correlation is not as strong as one might hope for. \added{A full replication would imply a correlation of 1, that is, all results being equal to the original study}. The detailed comparison for all data points is presented in Figure~\ref{fig:corr}.\footnote{\added{The B-cube splines smooth the regression line by using polynomial control points. This means that the regression line is allowed to adapt to the area of the data it is fitted to. The choice to use three splines is based on the number of expected control points. In our case, those are 1) negative values, 2) zero, 3) positive values.}} The direction of the fitted slope indicates that in general, the original results have higher effect sizes for the same data points than the new model. Given the introduced uncertainty through two additional regression parameters (areal and phylogenetic control), this is not too surprising. However, this also means that if we do not control for such effects and/or a small sample, we might over-estimate the true effect \parencite{Gelman2012, McElreath2020}.

\begin{figure}[ht]
    \centering
    \includegraphics[width=1\linewidth]{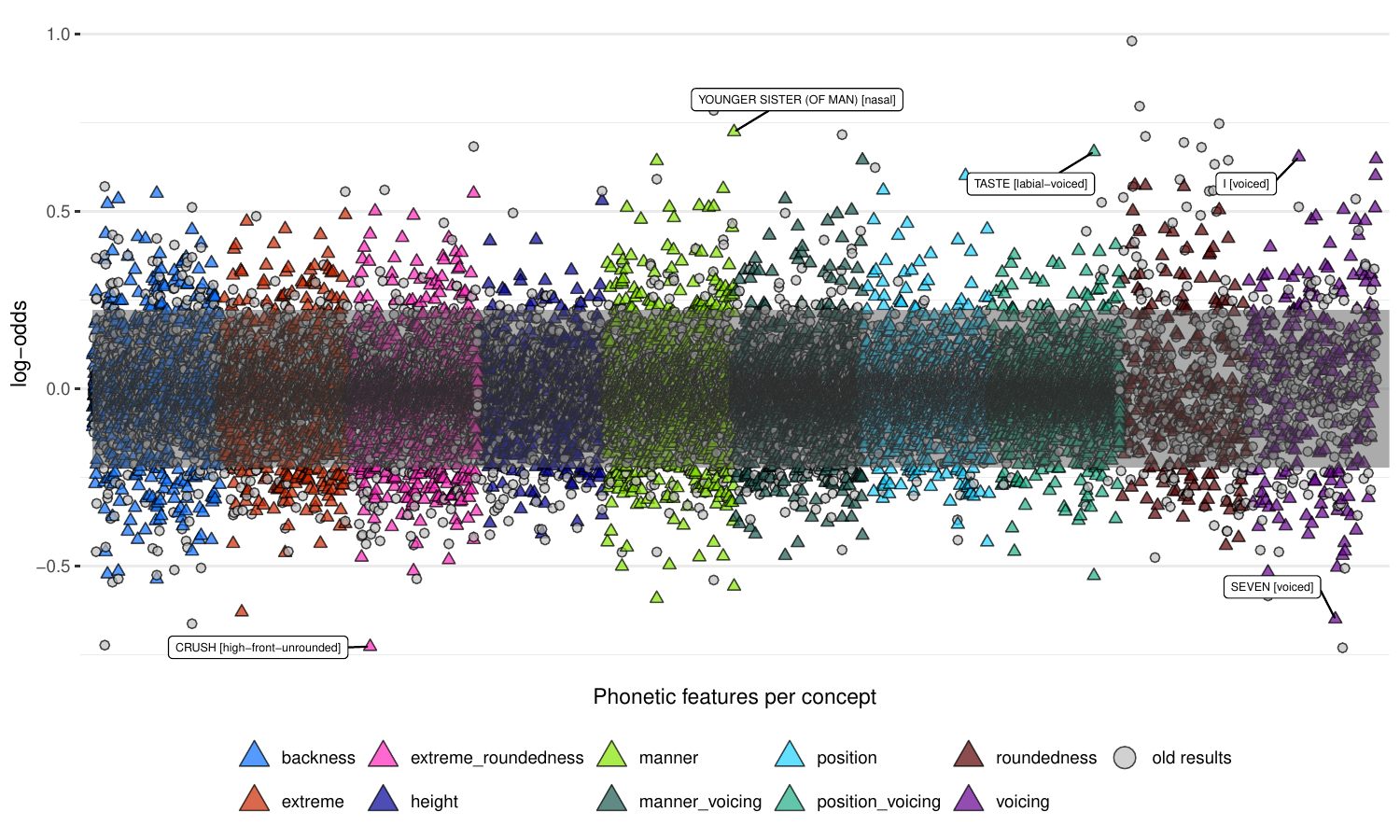}
    \caption{Comparison of old and new data mean effect sizes for all phonological categories. Grey \replaced{circles}{dots} indicate that the value is from the original results, while the colored \replaced{triangles}{dots} are the new results. The x-axis represents the individual phonological feature, grouped by category. The grouping is indicated by coloring. The shaded area around zero indicates the ROPE. \added{The five highest values of the new results are labeled.}} 
    \label{fig:manhattan}
\end{figure}

The summary Manhattan plot \parencite{Turner2018} in Figure~\ref{fig:manhattan} shows the distribution of effect sizes across the different concepts. \added{The values across models for each concept are presented on the same point at the x-axis. This makes a direct comparison between old and new results possible.} Again, we can observe that on average, the original model tends to have higher absolute means than the new model. On the other hand, there are many individual effects that have higher absolute means than in the original study, indicating diverging results. \added{We can also see that we find more meaningful effects in some phonological categories (e.g. extreme roundedness, manner) than in others (e.g. height, voicing).}

Overall, we find \replaced{fewer}{less} strong and weak effects across all categories compared to the original study. The total amount of effects observed are presented in Table~\ref{tab:results}. Crucially, we find way less strong effects (4) than in the original study (85), and also only about 24\% the amount of weak effects (105 compared to 437). The phonological category where most effects could be found relate to cardinal position (but not height) and rounding of vowels, as well as the manner and position of consonants, in combination with their voicing. Note that those numbers do not represent the amount of reproduced patterns, but only the raw amount of statistical over-representation that was observed in the new sample. Some of those represent new patterns that had not been observed previously.

\begin{figure}[ht]
    \centering
    \includegraphics[width=0.7\linewidth]{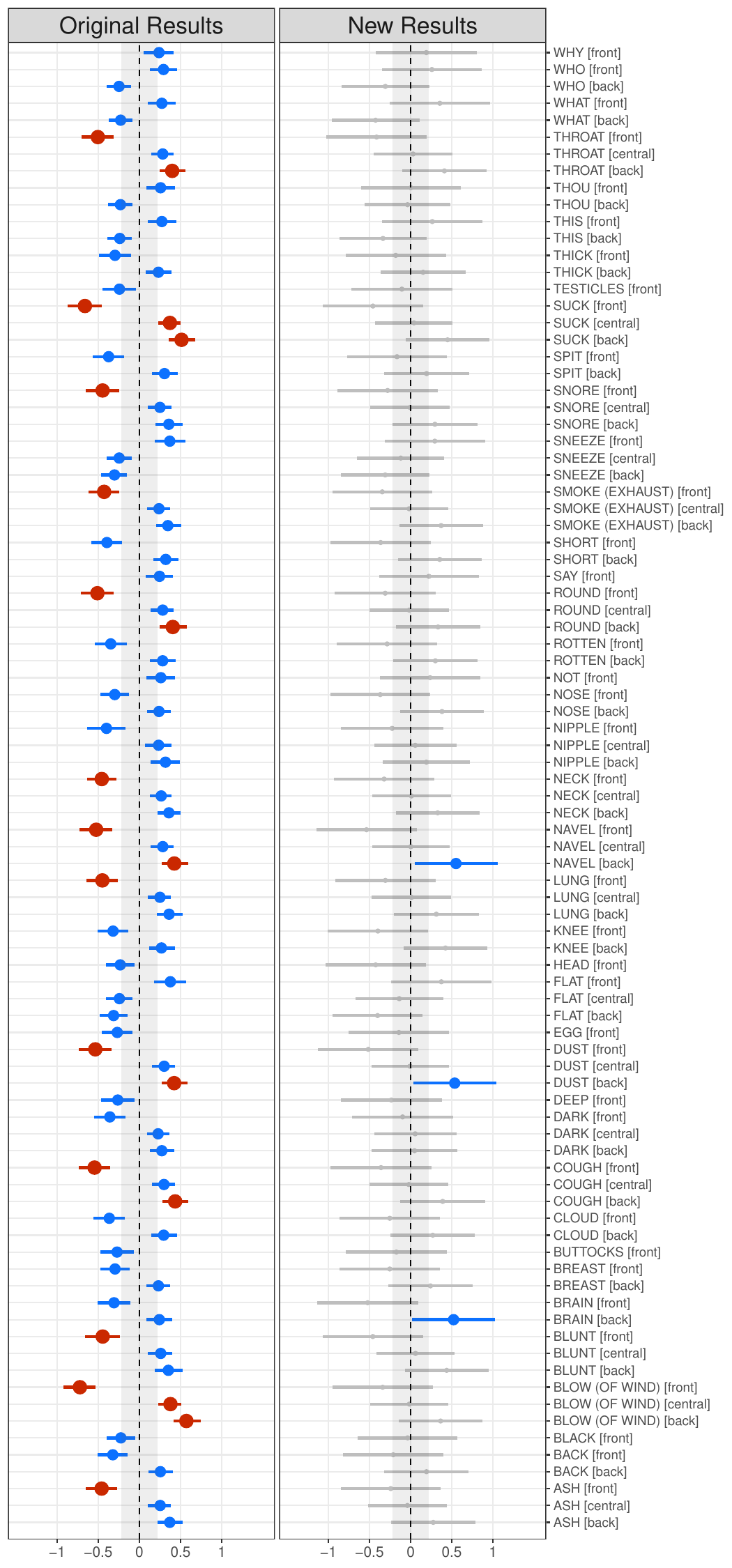}
    \caption{Comparison of results for `backness'. Red coloring indicates strong, and blue coloring weak results. Doubtful results are greyed out. The circle indicates the mean value, wheres the line shows the 95\% posterior distribution.}
    \label{res:backness}
\end{figure}

\deleted{From the comparisons in the Supplementary Material A (Section~\ref{supp_a}), we see that the posterior distributions for all parameters are much wider than in the original study.}
\added{
To dive deeper into the results, we present the posterior distribution of the model on `for all relevant concepts in Figure~\ref{res:backness}. We show the original and new results for all concepts that had strong (red) or weak (blue) evidence for an effect in either of the studies. The posterior distribution for the other nine phonological categories are presented in the Supplementary Material A (Section~\ref{supp_a})). In general, we can observe that, compared to the original analysis, only very few effects are robust. In many cases (e.g. \textsc{blow of wind [+central]}), the mean value of a previously strong effect is now almost 0. But there are also other cases where it is the wider 95\% HPDI (plotted line) which overlaps with the ROPE, despite the mean having a similar value (e.g. \textsc{round [+back]}). The increased uncertainty surrounding the draws from the posterior distributions are directly related to the added statistical controls, which make the parameter estimation for individual parameters more difficult \parencite{Gelman2012}. The complexity of our model results in more uncertainty in the inference, but is also less error prone \parencite{Becker2025}. This increased uncertainty helps us to avoid type 1 errors \parencite{Barr2013, Vasishth2016, Winter2021}, that is, \added{the probability of} reporting false positive effects. While the chosen value of the HPDI (in this case, 95\%) is to some kind arbitrary, it is exactly this kind of uncertainty analysis which allows us to identify the cases which are robust to a very high degree. The uncertainty directly reflects our trust in the statistical inference.
}

One category stands out with respect to the amount of observed patterns, namely the combination of vowel position and roundedness \added{(`extreme roundedness')}. All but one observations in this category are for the `high-back-rounded' feature \added{level (e.g. /u/)}. For the original results, two feature\added{ level}s were responsible for all the observations: `low-front-unrounded' \added{(e.g. /a/)}, and `high-front-unrounded' \added{(e.g. /i/)}. Interestingly, the first feature \added{level} can be considered diametrically opposite to the `high-back rounded' feature which we observe. While we have investigated this issue, we have found no annotations that could bias our model \replaced{towards}{into} this behavior. For example, the number of `high-back rounded' vowels is not higher than expected, and only features the third most prominent feature in the category, after `low-front unrounded' and `high-front-unrounded'. The fact that the previous study also has a strong imbalance with respect to which feature\added{s} present\deleted{s} effects, and which do not, makes us think that there might be some feature-specific reasons why those categories are over-represented, possibly related to the distinctiveness of phones in the vowel space.

Our results indicate that adding two additional controls reduces the number of observed effects. In total, we observe less than 21\% of the amount of previous effects. This points towards a stark over-estimation of potential patterns of sound symbolism in previous studies. Once we consider such dependencies as well as a \replaced{larger}{wider} sample of languages, most effects of statistical over-representation of phonological features disappear.\footnote{\added{We have also run models without removing the languages that formed part in the original study. In this case, we had even less concepts with evidence for strong and weak effects. We report on this in the Supplementary Material C.}} This has severe implications for the robustness of results that do not take such dependencies into account.

\begin{table}[t]
    \centering
    \begin{tabular}{|l|l|cc|cc|}
      \hline
        \rowcolor{gray!35}  \multicolumn{2}{|c|}{\textbf{Category}} &  \multicolumn{2}{|c|}{\textbf{Strong}} &  \multicolumn{2}{|c|}{\textbf{Weak}} \\  
        \rowcolor{gray!35} \multicolumn{2}{|c|}{} & Original & New & Original & New \\ \hline
        \multirow{5}*{Consonant features}   & voicing &   7 &   0 &  46 &   0 \\
                                   & manner &   9 &   1 &  52 &  18 \\ 
                                   & manner\_voicing &   6 &   0 &  53 &   8 \\ 
                                   & position &   2 &   0 &  11 &   7 \\ 
                                   & position\_voicing &   2 &   1 &  10 &  12 \\ \hline
        \multirow{5}*{Vowel features} & backness &  22 &   0 &  66 &   3 \\ 
                                   & extreme &   3 &   0 &  50 &   4 \\ 
                                   & extreme\_roundedness &  13 &   2 &  57 &  47 \\ 
                                   & height &   3 &   0 &  50 &   0 \\
                                   & roundedness &  18 &   0 &  42 &   6 \\ \hline
      & \textbf{Total} &  \textbf{85} &   \textbf{4} & \textbf{437} & \textbf{105} \\ 
      \hline
    \end{tabular}
    \caption{Number of effects found across all phonological categories.}
    \label{tab:results}
\end{table}

The four effects that come out as strong are for the concepts \textsc{taste} (labial-voiced), \textsc{dust} (high-back and rounded), \textsc{navel} (high-back rounded), and \textsc{younger sister (of man)} (nasal). While the original study found an effect for low-front and unrounded vowels with \textsc{taste}, the observation of an effect for position and voicing of consonants is new. At the same time, the observation of a higher likelihood for the vowel features could not be repeated. For \textsc{dust}, on the other hand, the original study found a negative effect for low-front and unrounded vowels, that is, those vowels being less likely to form part of the concept. The new results show a higher likelihood for high-back and rounded vowels instead. While they are not the same feature, they do complement each other. Similarly, \textsc{navel} was positively associated with a \textsc{+rounded} feature, while previously an effect was found for the combination of roundedness and a high-back position. The observation for \textsc{younger sister (of man)} is new and was not observed previously. In the remainder of this section, we will focus on the concepts that are part of the basic vocabulary lists. Of the strong results, only \textsc{navel} is part of one of the three lists of basic vocabulary we analyze here. A detailed discussion of further concepts will be given in the next section.

\subsection{Analyzing the previously observed patterns}
Of each of the previous findings within concepts of basic vocabulary, between 10\% and 25\% of the patterns could be reproduced. The quantitative results for all lists are presented in Table~\ref{tab:basic} and Table~\ref{tab:basic2}. For the Holman-40 and the the Swadesh-100 list, those concepts are \textsc{I} (nasal, stop) and \textsc{thou} (nasal). This shows that the strongest patterns of statistical over-representation of sounds regard the pronominal system. Both concepts are part of the three core basic vocabulary lists presented here. For the Tadmor-100 list, the previously observed patterns for the concepts \textsc{navel} (back vowel, rounded) and \textsc{suck} (rounded, high-back) could also successfully be replicated. All other observed patterns did not reproduce with the new sample. In conclusion, this means that only around 5\% of concepts in the three basic vocabulary \added{lists} are robustly affected by a statistical over-representation of certain phonological features - and those are cases which are well known among historical linguists and broadly discussed in the literature \parencite{nichols1996amerind, nichols2012, zamponi2017first}. Considering that we still found many effects in our analysis, this opens up the questions whether those would reproduce with new data and methodology. We will leave this discussion to further studies with other methodology and/or language samples.

\begin{table}[t]
    \centering
    \begin{tabular}{|l|c|cc|cc|} \hline
        \rowcolor{gray!35}  Vocabulary list & Original (n) &   New (n) &   New (\%) &   Replicated (n) &   Replicated (\%) \\ \hline
        \href{https://concepticon.clld.org/contributions/Swadesh-1955-100}{Swadesh-100} & 19 & 27 & 27\% & 2 & 10.5\% \\
        \href{https://concepticon.clld.org/contributions/Tadmor-2009-100}{Tadmor-100}  & 23 & 27 & 27\% & 4 & 17.4\% \\
        \href{https://concepticon.clld.org/contributions/Holman-2008-40}{Holman-40}  & 8 & 12 & 30\% & 2 & 25\% \\
    \hline \end{tabular}
    \caption{Number of concepts with over-represented phonological features in three lists of basic vocabulary.}
    \label{tab:basic}
\end{table}

\begin{table}[t]
    \begin{tabular}{|l|c|cc|cc|} \hline
         \rowcolor{gray!35} Vocabulary list & Original (n) &   New (n) &   New (\%) &   Replicated (n) &   Replicated (\%) \\ \hline
         \href{https://concepticon.clld.org/contributions/Swadesh-1955-100}{Swadesh-100} & 19 & 8 &     8\% & 2 & 10.5\% \\ 
         \href{https://concepticon.clld.org/contributions/Tadmor-2009-100}{Tadmor-100} & 23 & 13 &     13\% & 4 & 17.4\% \\
         \href{https://concepticon.clld.org/contributions/Holman-2008-40}{Holman-40} & 8 & 5 &     12.5\% & 2 & 25\% \\ \hline
    \end{tabular}
    \caption{Number of concepts with over-represented phonological features in three lists of basic vocabulary without taking into account the `extreme roundedness' feature.}
    \label{tab:basic2}
\end{table}

If we compare individual concepts with the results from other studies, a couple of relevant observations emerge. For example, an effect for the feature \textsc{+nasal} for \textsc{nose} could not be observed. The same \replaced{effect was shown to be absent}{lack of an effect has previously been reported} by \textcite{Cathcart2024}, who used a phylogenetic approach to investigate the evolution of such patterns over time. Similarly to our results and that of \textcite{ErbenJohansson2020}, they also report a strong effect of certain phones (/n/) for the pronominal concepts \textsc{I} and \textsc{we}. \replaced{In our study, this finding appears}{which we find} as effects for nasal and voicing. Going back to the original comparison that \textcite{ErbenJohansson2020} made based on their results and those of \textcite{Blasi2016}, those effects emerge as highly stable. Another effect found in multiple studies is the higher likelihood for a presence of /l/ for the concept of \textsc{tongue}. \textcite{ErbenJohansson2020} found an effect for \textsc{+alveolar, +voiced} (position), while our study found an effect for \textsc{+lateral, +voiced} (manner). Still, it is one of the few features where multiple similar effects could be observed. Therefore, those features emerge as the best candidates for a true statistical over-representation of sounds in semantic concepts.

The results show how we can leverage the availability of large-scale databases to test previous results in linguistic typology for robustness. By accessing the data with SQLite queries, we can easily access vast amounts of data. The available standardization in CLDF-datasets like Lexibank \parencite{Blum2025b} allows us to design complex queries that retrieve phonological and semantic information from reference catalogs like Glottolog \parencite{Glottolog}, Concepticon \parencite{Concepticon}, and CLTS \parencite{CLTS}. In this study, we could create the phylogenetic and areal covariance matrices through the information in Glottolog. By querying CLTS for all sounds in Lexibank, we annotated our data for the phonological categories as used in the original study. The queries are easily customizable for different features, and there is a lot of potential for using the same data for different types of replication and robustness studies. At the same time, the availability of such prior studies makes it possible to easily derive informative priors for the new studies \parencite{Gelman2012}.

\section{Conclusion}
We have compared the results from a previous study on the extent of the over-representation of phonological features in the basic vocabulary of languages of the world to a new analysis including a large sample of languages and new statistical controls. We find that most of this over-representation disappears when controlling for areal and phylogenetic effects, casting doubts on the robustness of previously observed effects. Of the effects previously observed for lists of basic vocabulary, we could only reproduce between 10\% and 25\%. Those were primarily the concepts related to the pronominal system (\textsc{1sg, 2sg}), or the body (\textsc{navel, suck, tongue}). All other observed patterns in the basic vocabulary lists did not reproduce. This indicates that the statistical over-representation of phonological features is less wide-spread than previously assumed, and does not generalize across the languages of the world. We argue that only those patterns should be considered as `true effects' that arise through different samples and methodologies. The concepts that we could successfully reproduce are thus good candidates to be true cases of over-represented phonological features.

Another perspective on the relevance of potentially sound symbolic concepts in basic vocabulary lists comes from \textcite{Cathcart2024}. Generally, they find that while the root-meaning traits can be conserved, `evolutionary mechanisms are often more likely to remove symbolic sounds than they are to create them' \parencite[1081]{Cathcart2024}. Overall, we can thus say that the influence of sound symbolism on lists of basic vocabulary is probably minimal, albeit there are individual concepts which need to be treated carefully. However, \added{in historical linguistics}, concepts like \textsc{suck}, \textsc{mother}, or \textsc{1sg} have been treated with special attention most of the time for this very reason \parencite{Anttila1972, Kaufman1995}.

While for the body-related terms an iconic relationship could be presented as a causal hypothesis for those patterns, the case of the pronominal system is more complex. However, it is barely a new finding; broad discussions in the literature around this topic exists, and it is generally assumed to be more closely tied to language evolution and phonological features than to a cognitive sound-meaning association \parencite{Traunmuller1994, gordon1995phonological, nichols1996amerind, nichols2012}. The other patterns, including those that we have observed in our new sample, must hold up to further tests with other languages and possibly improved statistical controls for further verification. Note that we are not saying that no onomatopoeic or iconic relationships between individual sounds and features exist. What we are looking for in this study\deleted{, however,} are \emph{global} patterns, that is, effects which are present in many different languages and families. We do not make any claims about specific words in individual languages.

Neither our nor previous results are the only source of truth, since any statistical model is always a simplification of the world \parencite{McElreath2020}. Making this uncertainty explicit helps us to accumulate and compare evidence of different sources, in order to arrive at conclusions that combine all types of available evidence. The results highlight the importance of robustness analysis for quantitative typology, in line with other recent calls for replication \parencite{Hartmann2024a, Becker2025}. This study also shows that we should not reduce our typological samples to a small number of languages, but rather try to take as much information as possible into account. When doing so,\deleted{however,} it is mandatory to include statistical controls for the various dependencies between languages. One of our goals should be to find the differences between the various ways of doing this, in order to find the adequate methods from a causal point of view \parencite{GuzmanNaranjo2023, verkerk_skirgard_2023}.

Robustness analyses provide an important assessment for testing the reliability of our scientific progress. For this purpose, it is indispensable to focus on reproducible science \parencite{Munaf2017}. We think that it is \replaced{crucial}{indispensable} to remain cautious with our analysis of statistical models and to employ sensible controls for areal and genealogical dependencies. To be able to do this, a commitment to sharing (when it is ethically justifiable) our code and data (like the original authors thankfully did) is an important step in this direction. We should embrace such contradictory results as signs of scientific progress \parencite{laudan1978progress, Ioannidis2012, Korbmacher2023} and find the best way to tackle those cases in discussion as well as collaboration. 

\section{Data Availability}
All code used for this study is curated on GitHub (+++link removed for anonymity+++). For the review, the code can be accessed anonymously on OSF: \url{https://osf.io/4kg52/?view_only=a96702c55db14528b9a3e7ed3701588b}
 
\printbibliography
\newpage
\pagenumbering{gobble}
\section{Supplementary Material A: Results of individual phonological categories}
\label{supp_a}

The following figures show the direct comparisons of original and new results for all ten phonological categories. We only show results that had observable effects in at least one of the studies. \added{The red coloring indicates strong results, and the blue coloring indicates weak results.}
 
\begin{figure}[ht]
    \centering
        \includegraphics[width=0.7\linewidth]{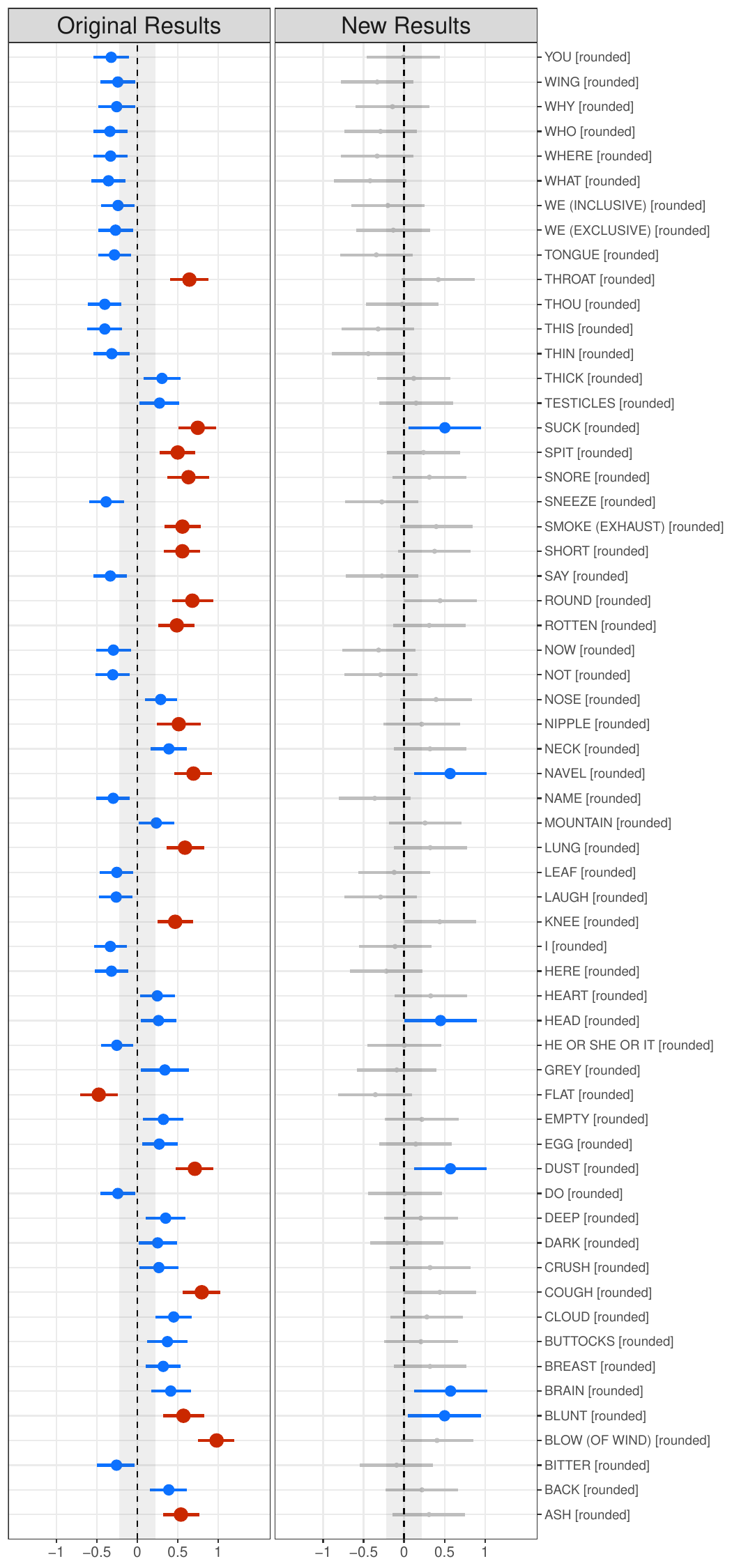}
        \caption{Comparison of results for `roundedness'.}
\end{figure}

\newgeometry{left=1cm, right=1cm}
\begin{figure}[ht]
    \centering
    \begin{subfigure}{.5\textwidth}
    \includegraphics[width=0.984\linewidth]{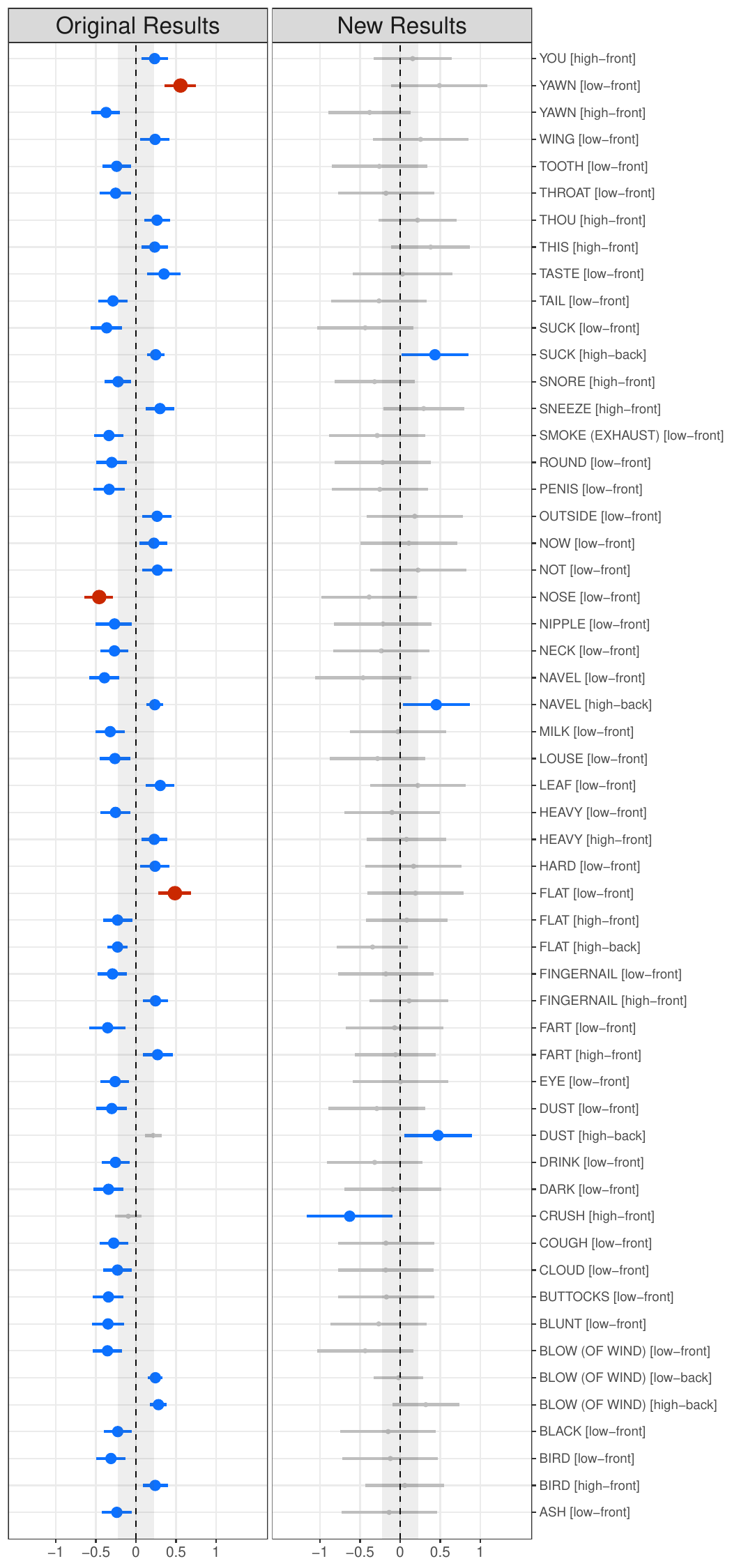}
    \caption{Comparison of results for `extreme'.}
    \end{subfigure}%
    \begin{subfigure}{.5\textwidth}
    \includegraphics[width=0.984\linewidth]{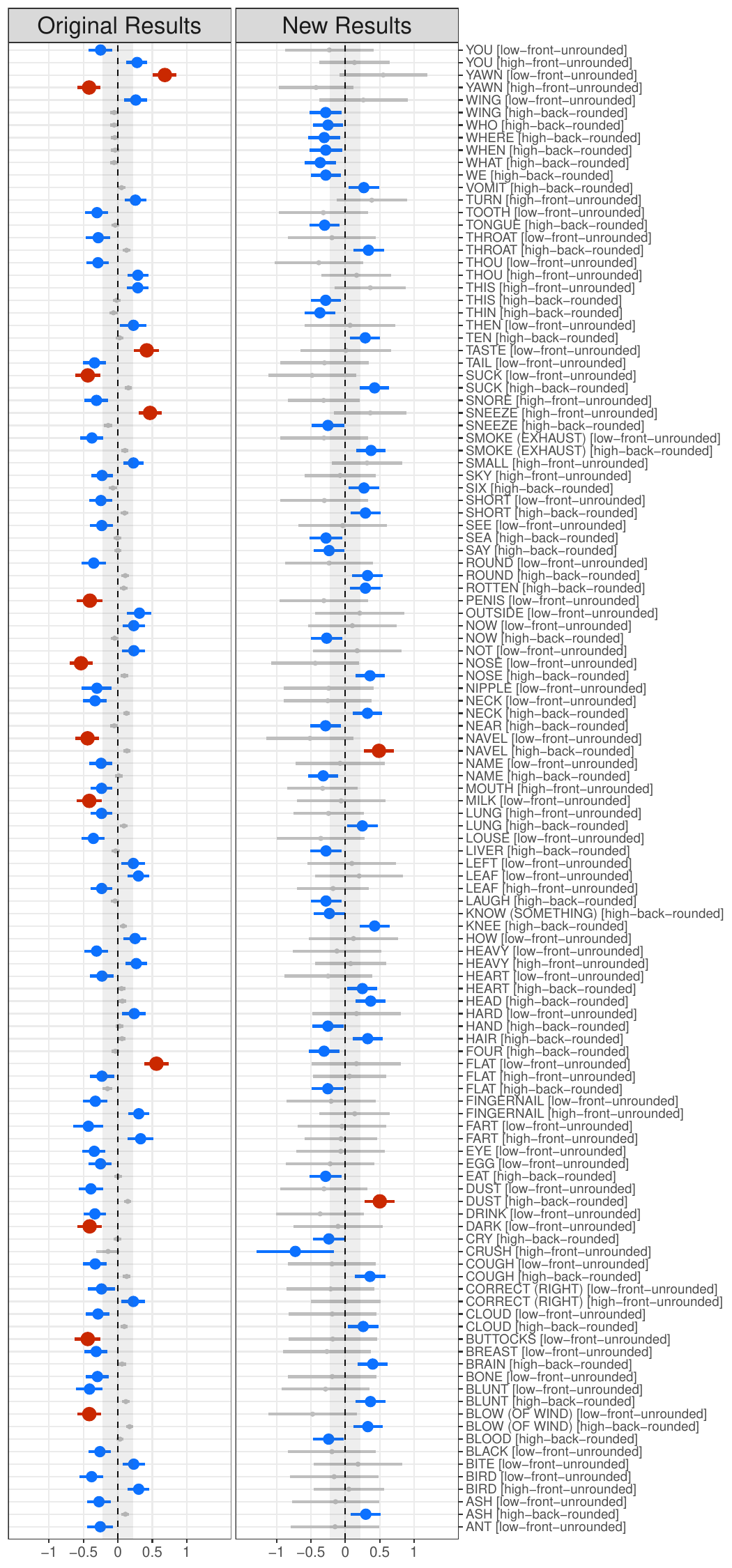}
    \caption{Comparison of results for `extreme roundedness'.}
    \end{subfigure}
\end{figure}

\newgeometry{left=1cm, right=1cm}
\begin{figure}[ht]
    \centering
    \begin{subfigure}{.5\textwidth}
    \includegraphics[width=0.984\linewidth]{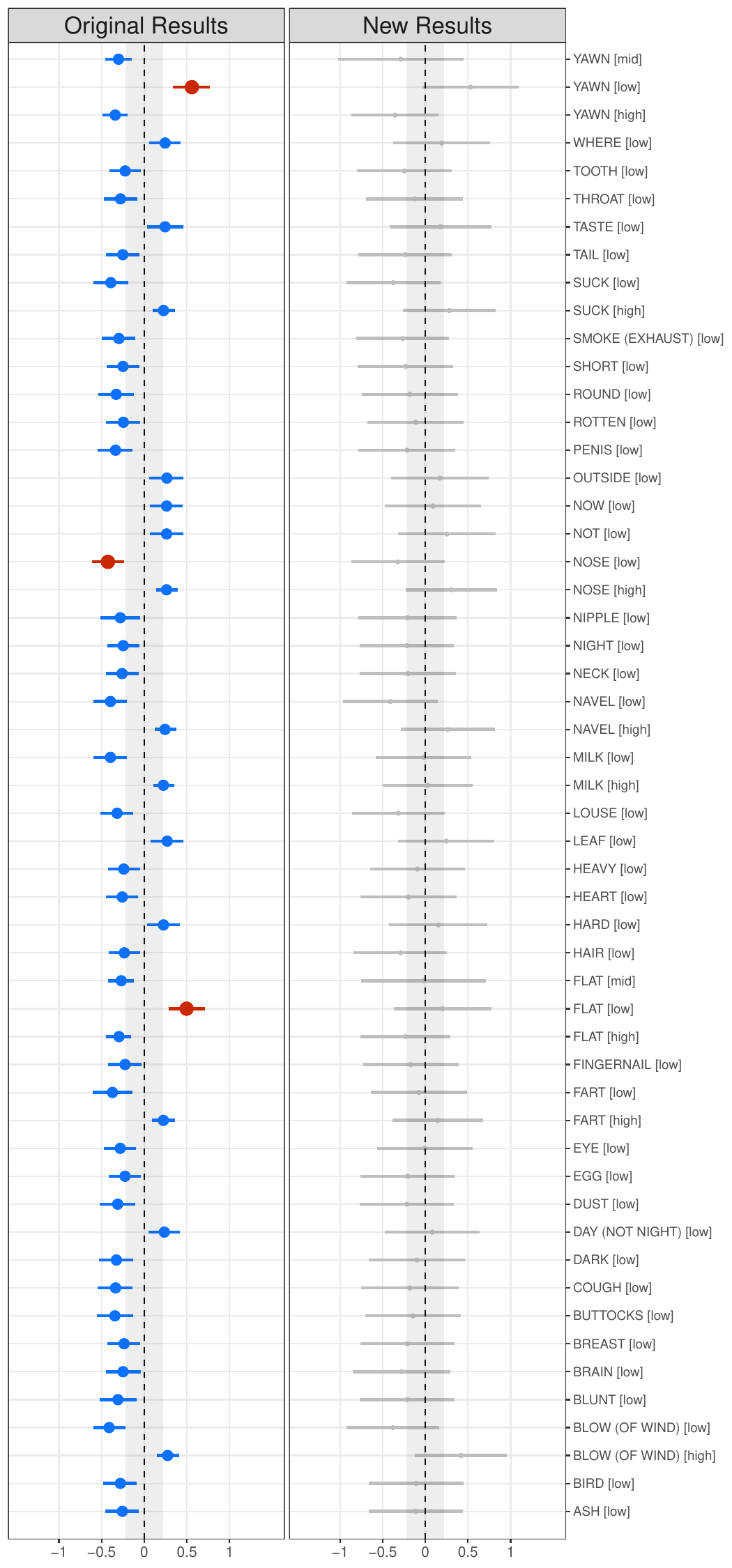}
    \caption{Comparison of results for `height'.}
    \end{subfigure}%
    \begin{subfigure}{.5\textwidth}
    \includegraphics[width=0.984\linewidth]{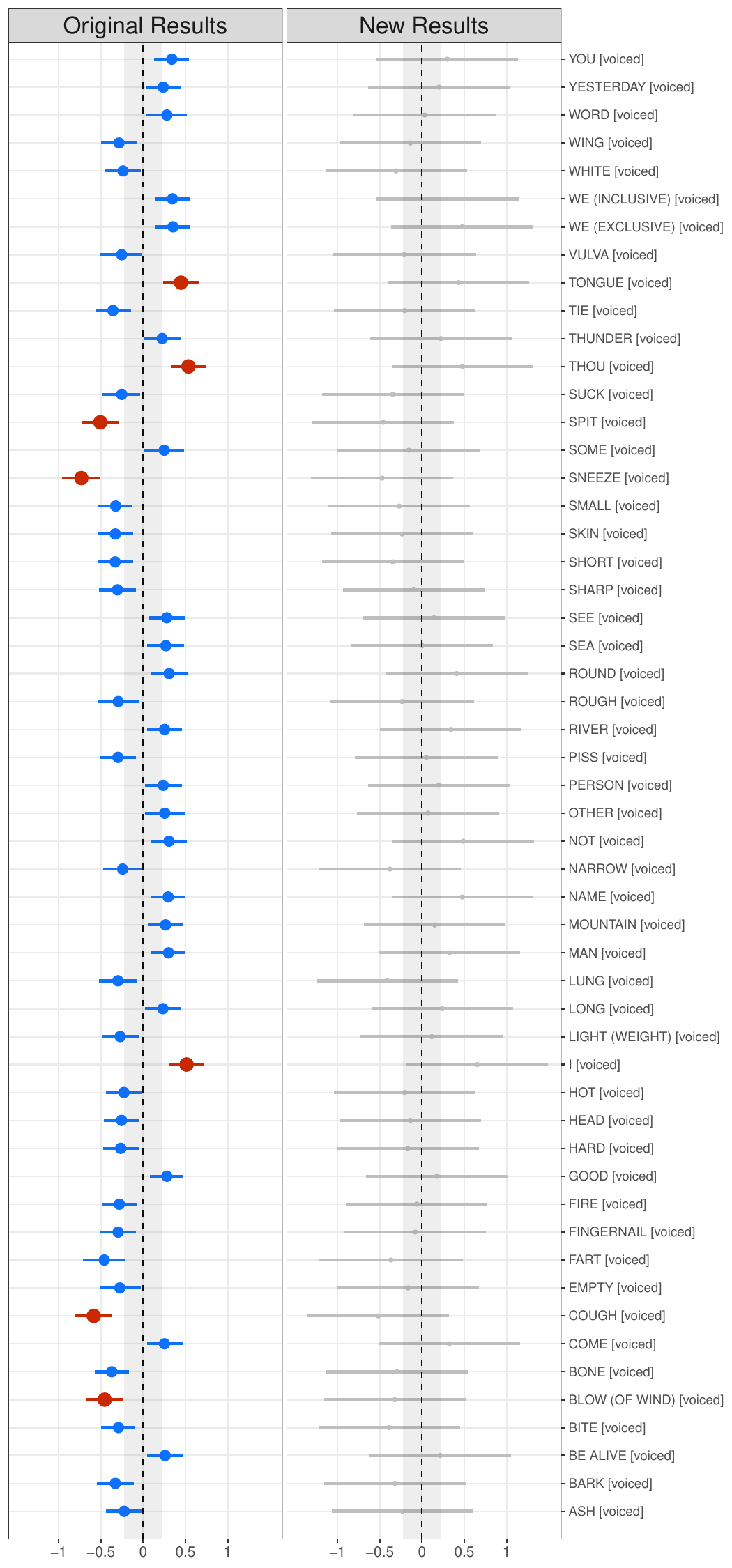}
    \caption{Comparison of results for `voicing'.}
    \end{subfigure}
\end{figure}

\newgeometry{left=1cm, right=1cm}
\begin{figure}[ht]
    \centering
    \begin{subfigure}{.5\textwidth}
    \includegraphics[width=0.984\linewidth]{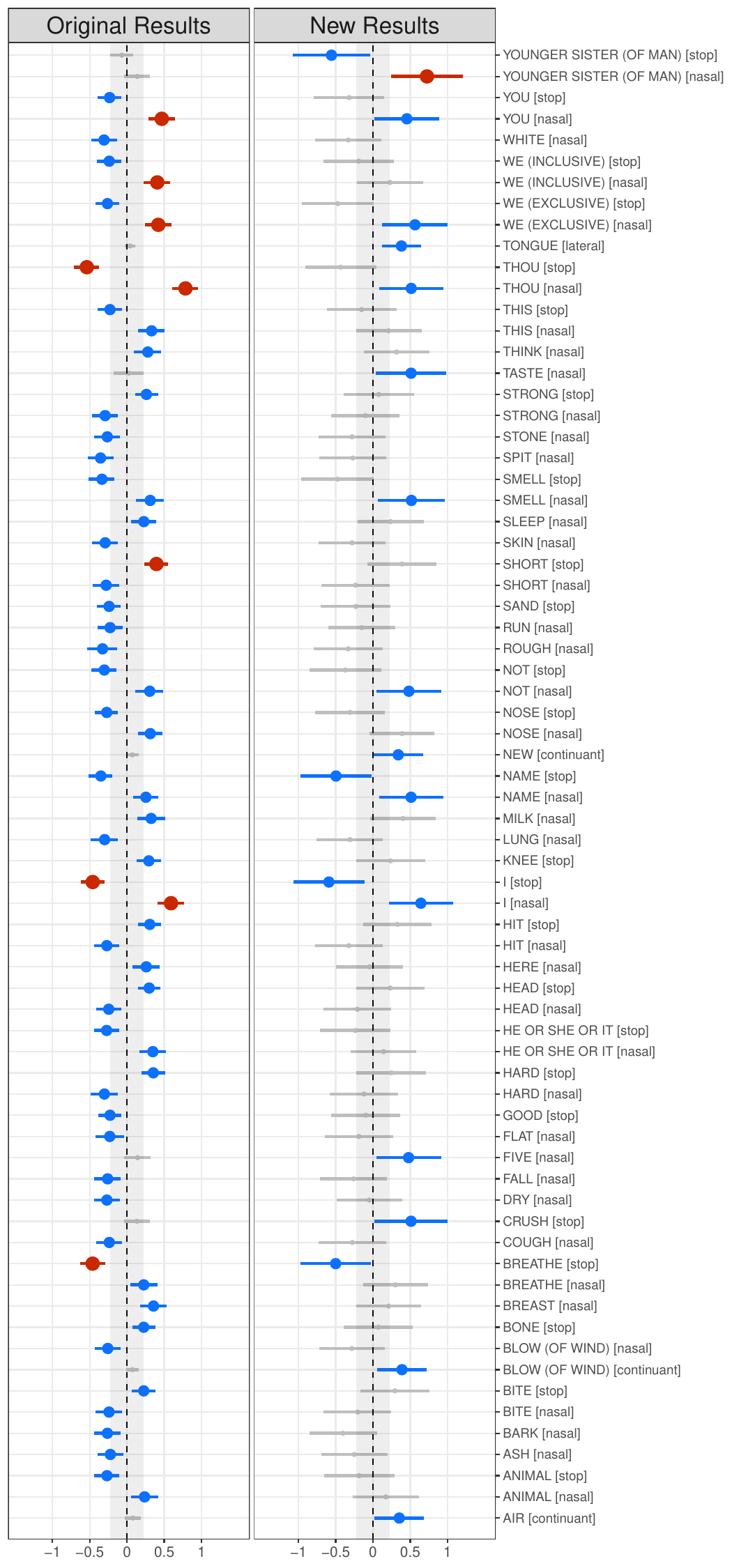}
    \caption{Comparison of results for `manner'.}
    \end{subfigure}%
    \begin{subfigure}{.5\textwidth}
    \includegraphics[width=0.984\linewidth]{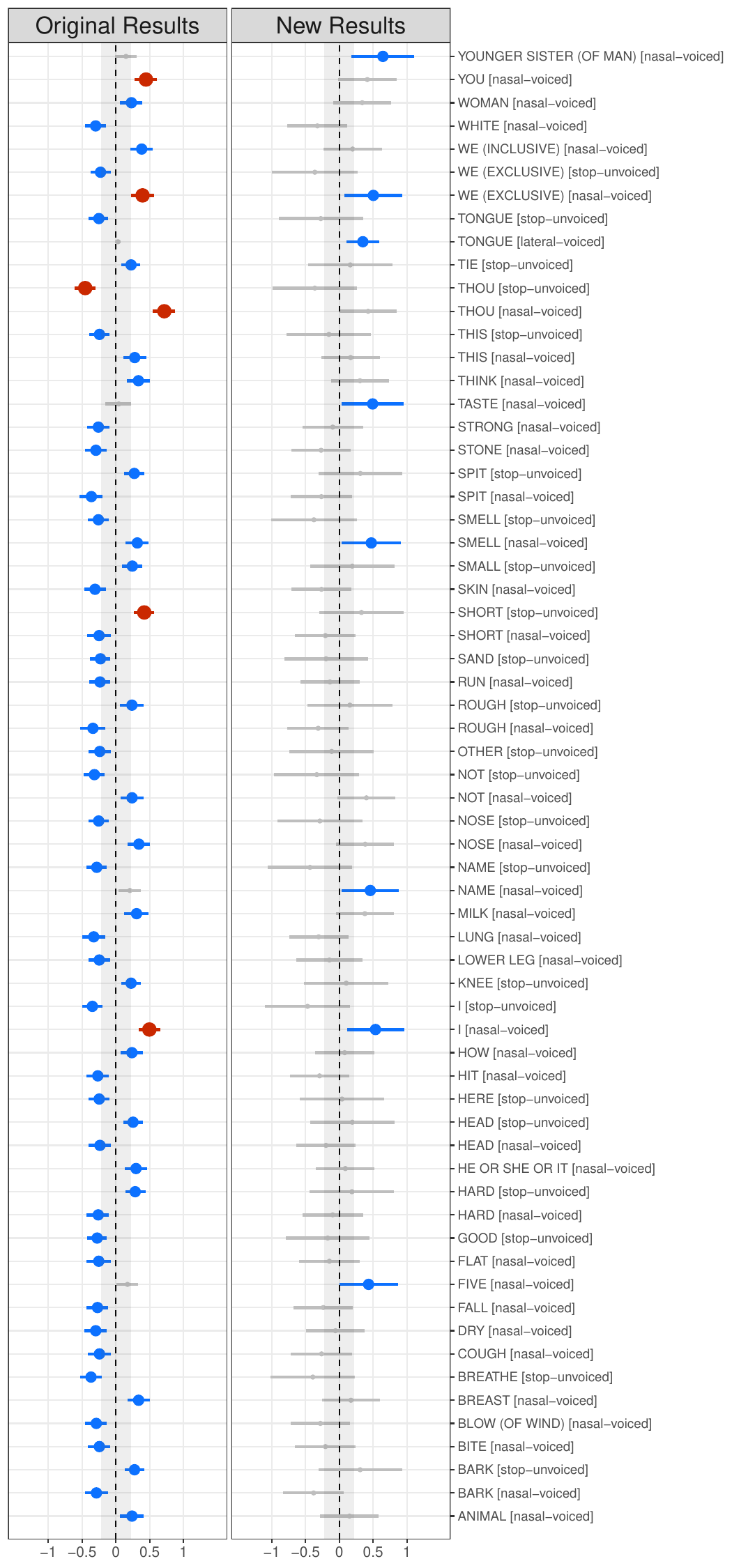}
    \caption{Comparison of results for `manner voicing'.}
    \end{subfigure}
\end{figure}

\newgeometry{left=1cm, right=1cm}
\begin{figure}[ht]
    \centering
    \begin{subfigure}{.5\textwidth}
    \includegraphics[width=0.984\linewidth]{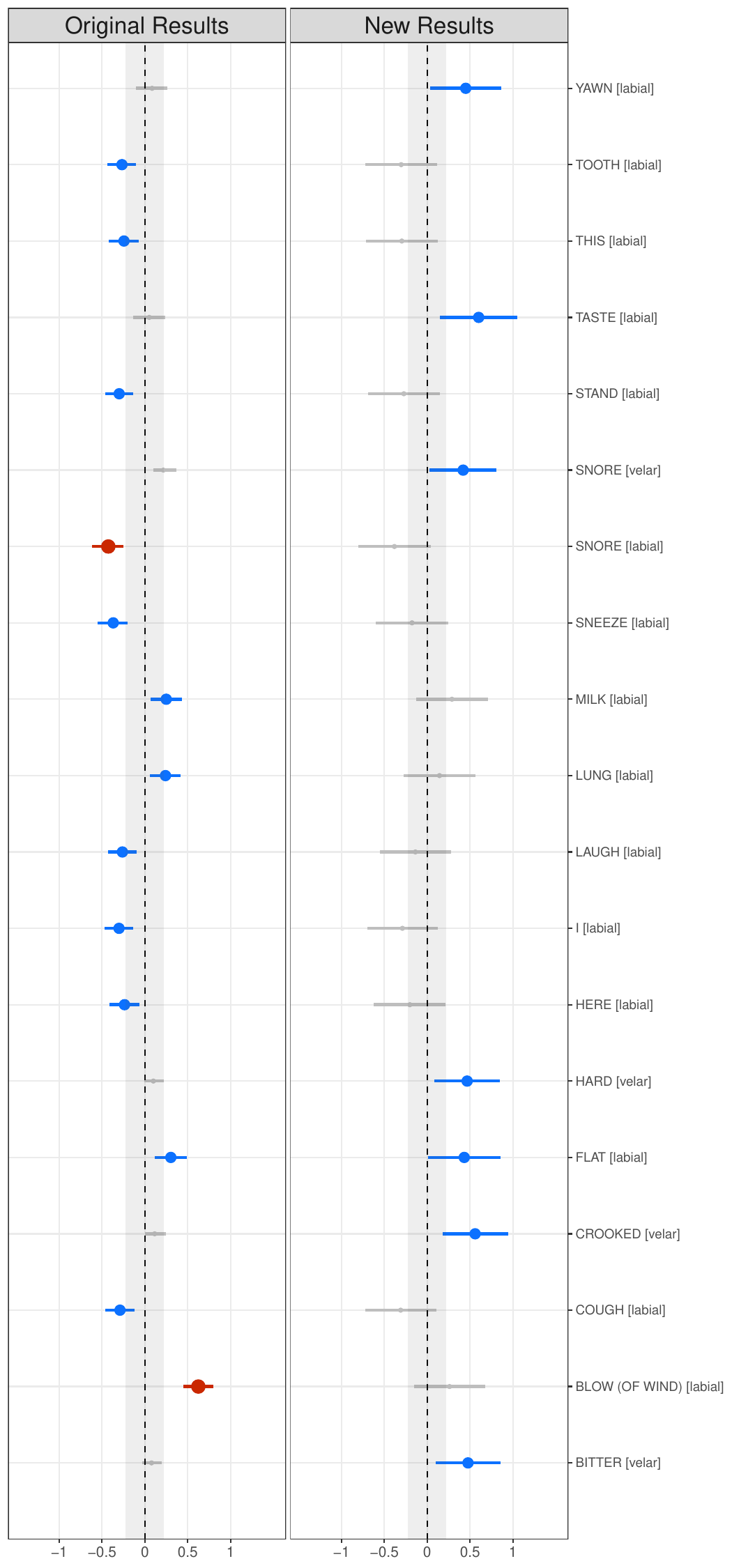}
    \caption{Comparison of results for `position'.}
    \end{subfigure}%
    \begin{subfigure}{.5\textwidth}
    \includegraphics[width=0.984\linewidth]{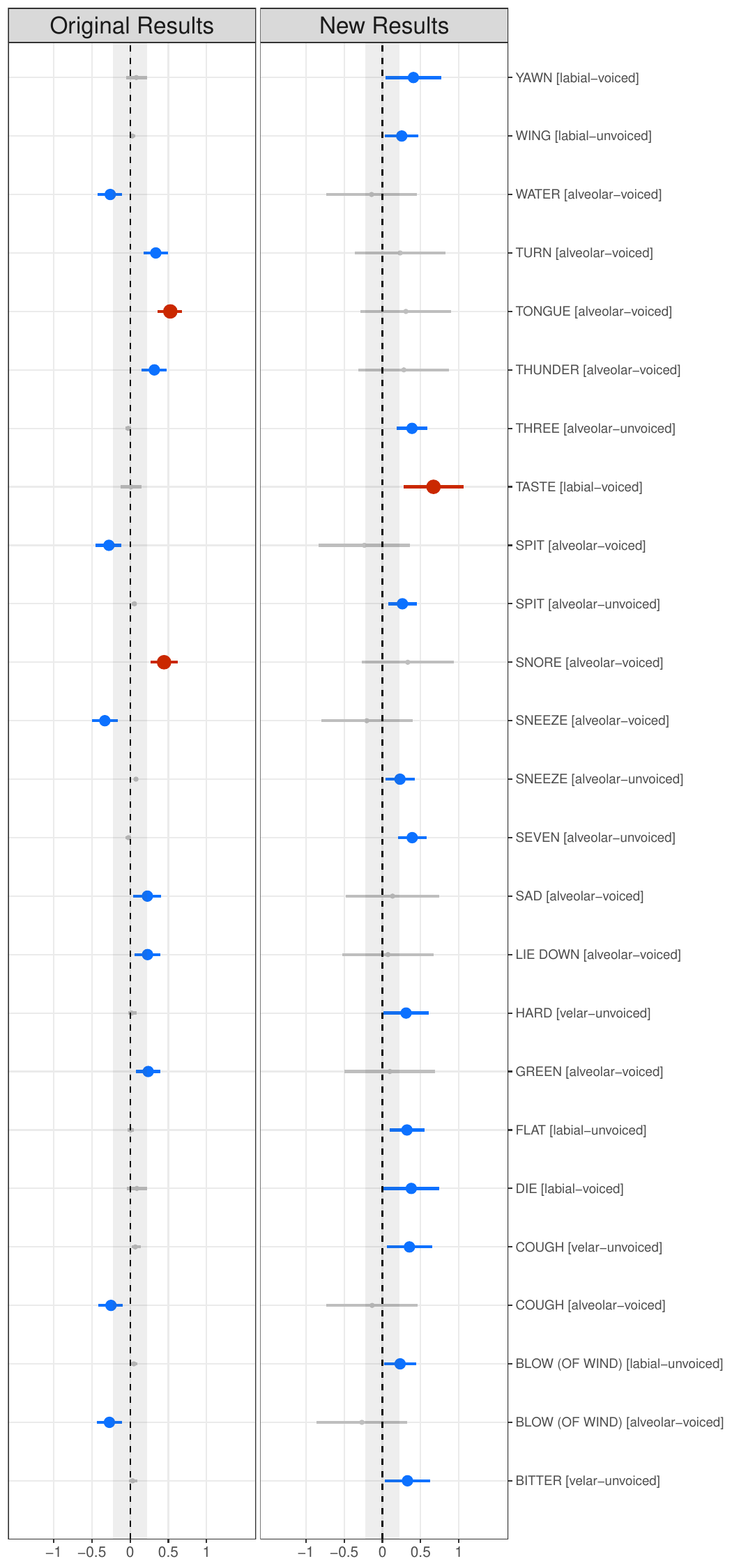}
    \caption{Comparison of results for `position voicing'.}
    \end{subfigure}
\end{figure}


\clearpage
\newgeometry{left=2cm, right=2cm}

\section{Supplementary Material B: Reproduction of previous results}
There were several problems when reproducing the number of sound symbolic concepts from the supplementary material of the original study. First, the results sheet uses different names from the concepts than the analysis-sheet for the number of affected concepts in the basic vocabulary lists. Thus, no automated matching is possible. A manual comparison for more than 300 concepts is not desirable. Second, it is unclear which threshold the original authors used. When using the thresholds that have been written out in the article (log(1.25)), we receive a different number of matches than the original study. Third, it is not always clear which concepts are actually compared. For example, the data distinguishes \textsc{we (exclusive)} from \textsc{we (inclusive)}. The basic vocabulary lists in question,\deleted{however,} do not make that distinction and have \textsc{we} (\url{https://concepticon.clld.org/values/Holman-2008-40-36}). The same issue arises e.g. with \textsc{blow (with mouth)} and \textsc{blow (of wind)}. The results file features an undifferentiated \textsc{blow}. In Concepticon, the entry in the Leipzig-Jakarta list is mapped to \textsc{blow (of wind)} (\url{https://concepticon.clld.org/values/Tadmor-2009-100-79}), in line with the original meaning from the World Loanword Database (\url{https://wold.clld.org/meaning/10-38#2/24.3/-4.8}). Given the sound symbolic interest,\deleted{however,} it is more likely that the authors were interested in \textsc{blow (with mouth)} and the associated round shape of the mouth. Such semantic ambiguities arise through a lack of standardization, which is a severe problem for the replication of such studies.

It was not clear to us how the authors of the original paper came to their numbers in Table 6 of the original paper. Loading their data and applying their proposed thresholds, we arrive at different number of matches. This matter gets complicated by the fact that the notion of 'semantic relatedness' is not made explicit. One could use the presence of colexifications for such relatedness \parencite{Blum2024d}, but it is not clear whether such a principled approach has been used here.

In the end, we decided to use the information provided in the Appendix A. Sadly, this sheet is not available in a machine-readable format. A conversion of the information is provided in the digitization to Concepticon (\url{https://concepticon.clld.org/contributions/Johansson-2020-344}). But ideally, we would have been able to reproduce this directly from the results-file instead. And even here, we find mismatches between data, concepts, and article. For example, \textsc{name} is missing from the list, but comes out clearly as sound symbolic in both the original study and our replication, with the same features as \textsc{1sg}. Interestingly, this seems to suggest a close relation between the first-person singular and the concept \textsc{name}, both having practically the very same phonological features from a statistical point of view. 

It is important for us to highlight that we do not intend this as scolding the authors of the original article. We could only find those problems because the authors have published most of their material as supplementary material. With most other studies, we wouldn't even have gotten so far.

\clearpage
\section*{Supplementary Material C: Model comparisons}

\subsection*{Excluding data from the original study}
\begin{figure}[th]
    \centering
    \includegraphics[width=1\linewidth]{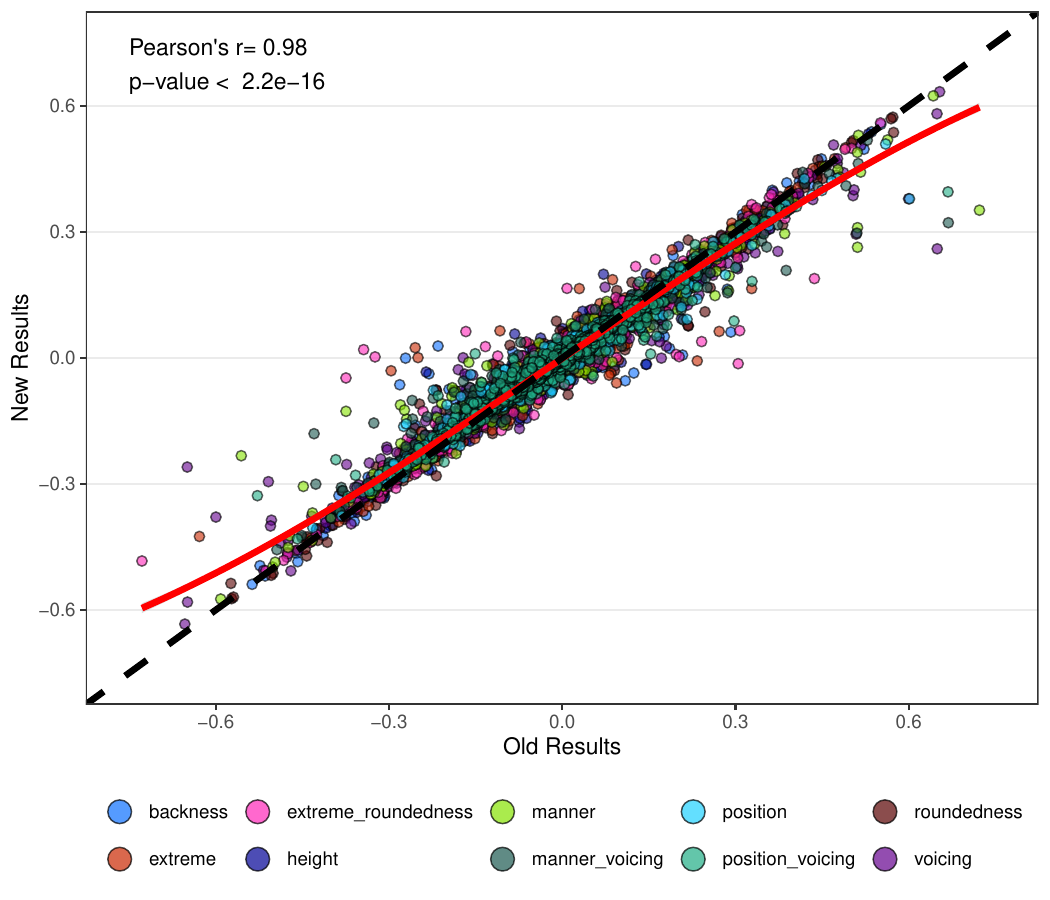}
    \caption{Posterior values from the full vs. the reduced models for all phonological variables.}
    \label{fig:full}
\end{figure}

\added{
In order to test the results on completely new data, we have removed data from all languages that have been used in the original study from our models. This raised the valid concerns from both anonymous reviewers that this might influence the results. To show that there is no strong bias resulting from this decision, we have run alternative models for all phonological features including data from all available languages. The results from the full vs. reduced models are strongly correlated (0.98), which shows that the differences between both are minimal. The correlation plot for both results are presented in Fig.~\ref{fig:full}.
}

\added{
We also have a numeric comparison of the strong and weak results. Interestingly, the full model shows even \emph{less} strong and weak results than the reduced sample. The full table is given in Table~\ref{tab:full}.
}

\begin{table}[ht]
    \centering
        \begin{tabular}{|l|rr|rr|}
          \hline
        \rowcolor{gray!35} Category & Reduced strong & Full strong & Reduced weak & Full weak \\ 
          \hline
          backness &   0 &   0 &   3 &   2 \\ \hline
          extreme &   0 &   0 &   4 &   3 \\ \hline
          extreme\_roundedness &   2 &   2 &  47 &  47 \\ \hline
          manner &   1 &   0 &  18 &   9 \\ \hline
          manner\_voicing &   0 &   0 &   7 &   5 \\ \hline
          position &   0 &   0 &   7 &   4 \\ \hline
          position\_voicing &   1 &   0 &  12 &   9 \\ \hline
          roundedness &   0 &   0 &  10 &  12 \\ \hline
          Total &   4 &   2 & 105 &  91 \\ \hline
        \end{tabular}
    \caption{Comparison of the replication with the reduced dataset and the full dataset.}
    \label{tab:full}
\end{table}

\subsection{Evaluating the predictive performance of models}
\added{
At request of a reviewer we also present the predictive performance of our sample via a leave-one-out cross validation \parencite{Vehtari2016}. Before turning to the results, it is important to highlight that the predictive performance does not need to equal truthfulness to the data generating process \parencite{Kuh2023, wang2014difficulty}. This means that we should not over-interpret the results of the model evaluation, and not select our preferred model based on the predictive performance but on causal reasoning.
}

\added{
For all ten phonological categories, we compared the full model (`with\_c'), a model with only a phylogenetic control (`phylo\_c'), a model with only areal control (`areal\_c'), and a model without multilevel controls (`no\_c'). In all cases, the model without multilevel controls performs worst. The model with areal controls performs better, but worse than the full and the phylogenetic model in all cases. This is similar than many other studies who used similar controls \parencite{Shcherbakova2023}. For most of the models, the full model has a slightly better predictive performance than the phylogenetic-only model, albeit with very small differences. The full results are presented in the Tables~\ref{first} to \ref{last}.
}

\begin{table}[ht]
\parbox{.32\linewidth}{
    \centering
    \begin{tabular}{|l|r|r|}\hline
    \rowcolor{gray!35} & \multicolumn{1}{c}{elpd\_diff} & \multicolumn{1}{c}{se\_diff} \\ \hline
    phylo\_c & 0.0 & 0.0 \\
    \hline
    with\_c & 2.7 & 3.0 \\
    \hline
    area\_c & 147.8 & 15.5 \\
    \hline
    no\_c & 18170.7 & 168.2 \\
    \hline
    \end{tabular}
    \caption{voicing}
    \label{first}
}
\parbox{.32\linewidth}{
    \centering
    \begin{tabular}{|l|r|r|}\hline
    \rowcolor{gray!35} & \multicolumn{1}{c}{elpd\_diff} & \multicolumn{1}{c}{se\_diff} \\ \hline
    with\_c & 0.0 & 0.0 \\
    \hline
    phylo\_c & 1.6 & 2.4 \\
    \hline
    area\_c & 159.0 & 13.0 \\
    \hline
    no\_c & 5264.0 & 79.4 \\
    \hline
    \end{tabular}
    \caption{roundedness}
}
\parbox{.32\linewidth}{
    \begin{tabular}{|l|r|r|}\hline
    \rowcolor{gray!35} & \multicolumn{1}{c}{elpd\_diff} & \multicolumn{1}{c}{se\_diff} \\ \hline
    phylo\_c & 0.0 & 0.0 \\
    \hline
    with\_c & 3.3 & 3.4 \\
    \hline
    area\_c & 208.0 & 18.6 \\
    \hline
    no\_c & 25004.1 & 160.3 \\
    \hline
    \end{tabular}
    \caption{height}
}
\end{table}

\begin{table}[ht]
\parbox{.32\linewidth}{
    \centering
    \begin{tabular}{|l|r|r|}\hline
    \rowcolor{gray!35} & \multicolumn{1}{c}{elpd\_diff} & \multicolumn{1}{c}{se\_diff} \\ \hline
    with\_c & 0.0 & 0.0 \\
    \hline
    phylo\_c & 2.8 & 3.2 \\
    \hline
    area\_c & 281.9 & 23.6 \\
    \hline
    no\_c & 19473.2 & 171.4 \\
    \hline
    \end{tabular}
    \caption{backness}
}
\parbox{.32\linewidth}{
    \centering
    \begin{tabular}{|l|r|r|}\hline
    \rowcolor{gray!35} & \multicolumn{1}{c}{elpd\_diff} & \multicolumn{1}{c}{se\_diff} \\ \hline
    phylo\_c & 0.0 & 0.0 \\
    \hline
    with\_c & 1.1 & 3.7 \\
    \hline
    area\_c & 428.5 & 23.4 \\
    \hline
    no\_c & 23913.5 & 168.5 \\
    \hline
    \end{tabular}
    \caption{extreme}
}
\parbox{.32\linewidth}{
    \centering
    \begin{tabular}{|l|r|r|}\hline
    \rowcolor{gray!35} & \multicolumn{1}{c}{elpd\_diff} & \multicolumn{1}{c}{se\_diff} \\ \hline
    phylo\_c & 0.0 & 0.0 \\ 
    \hline
    with\_c & 5.3 & 4.0 \\
    \hline
    area\_c & 508.0 & 31.3 \\
    \hline
    no\_c & 25055.1 & 173.6 \\
    \hline
    \end{tabular}
    \caption{position}
}
\end{table}

\begin{table}[ht]
\parbox{.32\linewidth}{
    \centering
    \begin{tabular}{|l|r|r|}\hline
    \rowcolor{gray!35} & \multicolumn{1}{c}{elpd\_diff} & \multicolumn{1}{c}{se\_diff} \\ \hline
    with\_c & 0.0 & 0.0 \\
    \hline
    phylo\_c & 4.7 & 4.1 \\
    \hline
    area\_c & 617.5 & 24.6 \\
    \hline
    no\_c & 20647.8 & 157.5 \\
    \hline
    \end{tabular}
    \caption{manner}
}
\parbox{.32\linewidth}{
    \centering
    \begin{tabular}{|l|r|r|}\hline
    \rowcolor{gray!35} & \multicolumn{1}{c}{elpd\_diff} & \multicolumn{1}{c}{se\_diff} \\ \hline
    with\_c & 0.0 & 0.0 \\
    \hline
    phylo\_c & 10.2 & 4.6 \\
    \hline
    area\_c & 1166.8 & 23.6 \\
    \hline
    no\_c & 38987.0 & 203.8 \\
    \hline
    \end{tabular}
    \caption{manner voicing}
}
\parbox{.32\linewidth}{
    \centering
    \begin{tabular}{|l|r|r|}\hline
    \rowcolor{gray!35} & \multicolumn{1}{c}{elpd\_diff} & \multicolumn{1}{c}{se\_diff} \\ \hline
    with\_c & 0.0 & 0.0 \\
    \hline
    phylo\_c & 4.3 & 4.5 \\
    \hline
    area\_c & 3089.1 & 44.3 \\
    \hline
    no\_c & 42249.7 & 246.2 \\
    \hline
    \end{tabular}
    \caption{extreme roundedness}
}
\end{table}

\begin{table}
    \centering
    \begin{tabular}{|l|r|r|}\hline
    \rowcolor{gray!35} & \multicolumn{1}{c}{elpd\_diff} & \multicolumn{1}{c}{se\_diff} \\ \hline
    with\_c & 0.0 & 0.0 \\ 
    \hline
    phylo\_c & 3.2 & 4.8 \\
    \hline
    area\_c & 2541.1 & 32.4 \\
    \hline
    no\_c & 38407.8 & 216.0 \\
    \hline
    \end{tabular}
    \caption{position voicing}
    \label{last}
\end{table}

\section*{Supplementary Material D: Phonological features}
\begin{table}[ht]
    \centering
    \begin{tabular}{|c|c|c|}
        \hline\rowcolor{gray!35}Feature & Levels & Example \\ \hline
        \multirow{2}*{roundedness} & rounded & /u/ \\
         & unrounded & /a/ \\ \hline
        \multirow{3}*{voicing}  & high & /i/ \\
         & low & /a/ \\
         & mid & /o/ \\ \hline
        \multirow{3}*{backness} & back & /u/ \\
         & central & /\textipa{@}/ \\
         & front & /a/ \\ \hline
        \multirow{4}*{extreme} & high-back & /u/ \\
         & high-front & /i/ \\
         & low-back & /\textipa{A}/ \\
         & low-front & /a/ \\ \hline
        \multirow{8}*{extreme roundedness} & high-back-unrounded & /\textipa{@}/ \\
         & high-front-unrounded & /i/ \\
         & low-back-unrounded & /\textipa{A}/ \\
         & low-back-rounded & /\textturnscripta/ \\
         & high-back-rounded & /u/ \\
         & high-front-rounded & /y/ \\
         & low-front-rounded & /\textipa{\ae}/ \\
         & low-front-unrounded & /a/ \\ \hline
    \end{tabular}
    \caption{Feature and levels for vowels. For `extreme roundedness', the features `roundedness' and `extreme' are combined. The example is the most frequent phoneme in the data.}
    \label{tab:vowels}
\end{table}

\begin{table}[ht]
    \centering
    \begin{tabular}{|c|c|c|}
        \hline\rowcolor{gray!35}Feature & Levels & Example \\ \hline
        \multirow{2}*{voicing}  & voicing & /n/ \\
                                & unvoiced & /k/ \\ \hline
        \multirow{5}*{position} & alveolar & /n/ \\
                                & glottal & /h/ \\
                                & labial & /m/ \\
                                & palatal & /j/ \\
                                & velar & /k/ \\ \hline
        \multirow{5}*{manner}   & continuant & /s/ \\
                                & lateral & /l/ \\
                                & nasal & /n/ \\
                                & stop & /k/ \\
                                & vibrant & /r/ \\ \hline
        \multirow{7}*{manner voicing}   & continuant unvoiced & /s/ \\
                                        & continuant voiced & /j/ \\
                                        & lateral voiced & /l/ \\
                                        & nasal voiced & /n/ \\
                                        & stop unvoiced & /k/ \\
                                        & stop voiced & /b/ \\
                                        & vibrant voiced& /r/ \\ \hline
        \multirow{10}*{position voicing} & alveolar voiced & /n/ \\
                                        & alveolar unvoiced & /t/ \\
                                        & labial voiced & /m/ \\
                                        & labial unvoiced & /p/ \\
                                        & velar unvoiced & /k/ \\
                                        & velar voiced & /g/ \\
                                        & palatal voiced & /j/ \\
                                        & palatal unvoiced & /c/ \\
                                        & glottal unvoiced & /h/ \\
                                        & glottal voiced & /\textipa{Q}/ \\\hline
    \end{tabular}
    \caption{Features and levels for consonants. For `manner voicing' and `position voicing', the features `manner' and `position' are combined with . The example is the most frequent phoneme in the data.}
    \label{tab:consonants}
\end{table}

\end{document}